%% file: ms.tex
\documentclass[10pt,twocolumn,letterpaper]{article}
\pdfoutput=1

\usepackage{iccv}
\usepackage{lmodern}
\usepackage{times}
\usepackage{epsfig}
\usepackage{graphicx}

\usepackage{amsmath}
\usepackage{amssymb}
\usepackage{bm}  
\usepackage{xfrac}
\usepackage{siunitx}
\usepackage{array}
\usepackage{tabularx}
\usepackage{booktabs}
\usepackage{makecell}
\usepackage{multirow}

\usepackage[labelsep=period]{caption}
\usepackage{subcaption}
\usepackage[inline]{enumitem}
\usepackage{lipsum}
\usepackage{placeins}  

\makeatletter
\renewcommand{\paragraph}{%
	\@startsection{paragraph}{4}%
	{\z@}{10pt \@plus 1ex \@minus .2ex}{-1em}%
	{\normalfont\normalsize\bfseries}%
}
\makeatother

\usepackage{tikz}
\usepackage{tikzscale}
\usepackage{pgfplots}
\usepgfplotslibrary{groupplots}
\usetikzlibrary{shapes, external, shadings, calc, patterns}
\input{settings/tikz}

\input{settings/commands}
\input{settings/glossary}
\input{figures/data/data-commands}

\usepackage[
	pagebackref=true,  
	breaklinks=true,colorlinks,bookmarks=false]{hyperref}

\usepackage[capitalize]{cleveref}  
\newcommand{\crefrangeconjunction}{\,--\,}
\crefname{table}{Tab.}{Table}

\iccvfinalcopy 

\def\iccvPaperID{19} 

\ificcvfinal\fi

\newcommand{\papertitle}{%
	LORD: Leveraging Open-Set Recognition with Unknown Data
}

\newcommand{\authors}{%
	\author{Tobias Koch\myMark{{*, \textdagger}}, Christian Riess\myMark{\textdagger}, and Thomas Köhler\myMark{*}\\
	\myMark{*}e.solutions GmbH, Erlangen, Germany\\
	\myMark{\textdagger}Friedrich-Alexander-Universität Erlangen-Nürnberg, Erlangen, Germany\\
	{\tt\small \{tobias.koch, thomas.koehler\}@esolutions.de, christian.riess@fau.de}}
}

\newcommand{\placetextbox}[3]{
	\setbox0=\hbox{#3}
	\AddToShipoutPicture*{
		\put(\LenToUnit{#1\paperwidth},\LenToUnit{#2\paperheight}){\vtop{{\null}\makebox[0pt][c]{#3}}}%
	}%
}%
\newcommand{\copyrightnotice}{%
		{\footnotesize%
	\begin{minipage}{\textwidth}%
		Accepted at ICCVW 2023.\\
		\textcopyright\ 2023 IEEE. Personal use of this material is permitted. Permission from IEEE must be obtained for all other uses, in any current or future media, including reprinting/republishing this material for advertising or promotional purposes, creating new collective works, for resale or redistribution to servers or lists, or reuse of any copyrighted component of this work in other works.
	\end{minipage}}%
}

\begin{document}

	\input{00paper}
	\clearpage
	\input{10supmat}

\end{document}

%% file: settings/tikz.tex
\pgfplotsset{
	compat=1.16,
}

\newcommand{\tikzref}[1]{%
	\tikzsetnextfilename{#1}%
	\ref*{#1}%
}

\newcommand{\includegraphicstikz}[2]{%
	\tikzsetnextfilename{#1}%
	\includegraphics[width=#2]{figures/#1.tikz}%
}

\definecolor{color-lb}{rgb}{.255, .714, .769}  
\definecolor{color-purple}{rgb}{0.533, .337, .855}  
\definecolor{color-green}{rgb}{.192, .639, .329}  
\definecolor{color-blue}{rgb}{0., 0., .55}  
\definecolor{color-orange}{rgb}{1., .55, 0.}  
\definecolor{color-darkgrey}{rgb}{.55, .55, .55}  
\definecolor{color-html-gray}{HTML}{808080}  
\definecolor{my-green}{HTML}{92d3bf}  
\definecolor{my-red}{HTML}{fc9369}  
\definecolor{my-green-four}{HTML}{3a9276}  
\definecolor{my-red-four}{HTML}{c83c04}  

\definecolor{blind-green}{HTML}{21c493}  
\definecolor{blind-orange}{HTML}{fc6f03}  
\definecolor{blind-violet}{HTML}{413e75}  

\colorlet{color-bl}{color-html-gray}
\colorlet{color-spl}{blind-orange}
\colorlet{color-mpl}{blind-green}
\colorlet{color-kvr}{blind-violet}

\newcommand{\tikzuuc}[1]{%
	\tikzsetnextfilename{#1}%
	\begin{tikzpicture}[baseline=-1mm]
		\node[star, draw=black, fill=black, scale=.65] (s) at (0,0) {};
	\end{tikzpicture}%
}

\newcommand{\tikzkuc}[1]{%
	\tikzsetnextfilename{#1}%
	\begin{tikzpicture}[baseline=-1mm]
		\node[regular polygon, regular polygon sides=6, fill=color-purple, scale=.75] (c) at (0,0) {};
	\end{tikzpicture}%
}

\newcommand{\mixup}{cnt_mixup_relative}

\pgfplotsset{short-legend-refs/.style={
	legend image code/.code={
		\draw[mark repeat=2,mark phase=2]
		plot coordinates {
			(0cm,0cm)
			(0.15cm,0cm)        
			(0.3cm,0cm)         
		};%
	},
}}

\pgfplotsset{
	log x ticks with fixed point/.style={
		xticklabel={
			\pgfkeys{/pgf/fpu=true}
			\pgfmathparse{exp(\tick)}%
			\pgfmathprintnumber[fixed relative, precision=3]{\pgfmathresult}
			\pgfkeys{/pgf/fpu=false}
		}
	},
	log y ticks with fixed point/.style={
		yticklabel={
			\pgfkeys{/pgf/fpu=true}
			\pgfmathparse{exp(\tick)}%
			\pgfmathprintnumber[fixed relative, precision=3]{\pgfmathresult}
			\pgfkeys{/pgf/fpu=false}
		}
	}
}

\tikzset{aucnodestyle/.style={
	draw,
	fill=white,
	align=left,
	font=\tiny,
	anchor=south east,
}}

\tikzset{legendnodestyle/.style={
	draw,
	align=left,
	font=\footnotesize,
	inner sep=1pt,
	anchor=south,
}}

\tikzset{subcaptionnodestyle/.style={
	text=black,
	align=center,
	font=\small,
	inner sep=0pt,
	outer sep=0pt,
	anchor=north,
}}

\pgfplotsset{my-basestyle/.style={
	text=black,
	tick style={black},
	axis line style={black},
	title style={font=\footnotesize},
	label style={font=\footnotesize},
	tick label style={font=\footnotesize},
	legend style={font=\footnotesize,text=black},
	legend cell align=left,
	enlarge x limits={abs=0.5em},
	enlarge y limits={abs=1em},
	no markers,
	grid=major,
	title style={yshift=-0.5ex,},
}}

\pgfplotsset{my-basestyle-curve/.style={
	my-basestyle,
	log x ticks with fixed point,
	every axis plot/.append style={line width=1.6pt},
}}

\pgfplotsset{my-basestyle-group/.style={
	my-basestyle,
	height=6em,
	width=\textwidth/3.0-2em,
	log x ticks with fixed point,
}}

\pgfplotsset{my-basestyle-curve-roc/.style={
	my-basestyle-curve,
	ylabel={TPR [\si{\percent}]},
	xlabel={FPR [\si{\percent}]},
	xlabel style={yshift=0.5ex},
	ymin=0,
	ytick={0,20,...,100},
}}

\pgfplotsset{my-basestyle-curve-roc-uu/.style={
	my-basestyle-curve-roc,
	xlabel={FPR$_{UU}$ [\si{\percent}]},
}}

\pgfplotsset{my-basestyle-curve-ccr/.style={
	my-basestyle-curve,
	ylabel={CCR [\si{\percent}]},
	xlabel={FPR [\si{\percent}]},
	xlabel style={yshift=0.5ex},
	ymin=0,
	ytick={0,20,...,100},
	restrict expr to domain={rawx}{:10},
}}

\pgfplotsset{my-basestyle-curve-ccr-genuine-plots/.style={
	my-basestyle-curve-ccr,
	enlarge y limits={abs=.5em},
}}

\pgfplotsset{my-basestyle-curve-ccr-uu/.style={
	my-basestyle-curve-ccr,
	ylabel={CCR [\si{\percent}]},
	xlabel={FPR$_{UU}$ [\si{\percent}]},
}}

\pgfplotsset{my-basestyle-curve-mixup/.style={
	my-basestyle-curve,
	xlabel={$\#$ \MixKnownRatio},
	xlabel style={yshift=0.5ex},
}}

\pgfplotsset{my-basestyle-curve-mixup-aucroc/.style={
	my-basestyle-curve-mixup,
	ylabel={\acrshort{aucroc} [\si{\percent}]},
}}

\pgfplotsset{my-basestyle-curve-mixup-ccr-at-fpr/.style={
	my-basestyle-curve-mixup,
}}

\pgfplotsset{my-basestyle-curve-cutout/.style={
	my-basestyle,
	enlarge x limits={abs=.25em},
	enlarge y limits={abs=.25em},
	axis background/.style={fill=white},
	ticklabel style={fill=white,inner sep=0pt,outer sep=0pt},
	xticklabel style={yshift=-1pt},
	yticklabel style={xshift=-1pt},
	every axis plot/.append style={line width=0.9pt},
	tiny,
	height=3em,
	width=4.5em,
}}

\pgfplotsset{style-bl/.style={
	solid,
	color=color-bl,
}}

\pgfplotsset{style-kvr/.style={
	solid,
	color=color-kvr,
}}

\pgfplotsset{style-spl/.style={
	solid,
	color=color-spl,
}}

\pgfplotsset{style-mpl/.style={
	solid,
	color=color-mpl,
}}

\pgfplotsset{style-lb/.style={
	solid,
	color=color-lb,
}}

\pgfplotsset{style-limit/.style={
	dashed,
	color=black,
}}

\pgfplotsset{style-kvr-dashed/.style={
	style-kvr,
	dashed,
}}

\pgfplotsset{style-kvr-dotted/.style={
	style-kvr,
	dotted,
}}

\pgfplotsset{style-spl-dashed/.style={
	style-spl,
	dashed,
}}

\pgfplotsset{style-mpl-dashed/.style={
	style-mpl,
	dashed,
}}

\pgfplotscreateplotcyclelist{list-bl-3-limit}{
	style-bl\\
	style-spl\\
	style-mpl-dashed\\
	style-kvr-dotted\\
	style-limit\\
}

\pgfplotscreateplotcyclelist{list-bl-4-limit}{
	style-bl\\
	style-spl\\
	style-lb\\
	style-mpl-dashed\\
	style-kvr-dotted\\
	style-limit\\
}

\pgfmathsetlengthmacro\BarWidth{5pt}
\pgfplotsset{my-basestyle-barplot/.style={
	my-basestyle,
	xtick={1,2,3,4,5,6},
	xticklabels={OSNN, DNN, EVM, C-EVM, W-SVM, \acrshort{pisvm}},
	ybar=0.5pt,
	bar width=\BarWidth,
	ylabel style={align=center,xshift=-2ex},
	ylabel={Gain in \ccrfpr{1}},
	xmajorgrids=false,
	enlarge x limits={abs=2.2*\BarWidth},
	enlarge y limits={abs=.5em},
	xtick style={draw=none},
	xticklabel style={yshift=6pt,font=\scriptsize},
}}

%% file: settings/commands.tex
\makeatletter

\DeclareRobustCommand\onedot{\futurelet\@let@token\@onedot}
\def\@onedot{\ifx\@let@token.\else.\null\fi\xspace}

\def\eg{\emph{e.\,g}\onedot} 
\def\ie{\emph{i.\,e}\onedot} 
\def\cf{\emph{cf}\onedot} 
 \def\vs{\emph{vs}\onedot}
\def\wrt{w.\,r.\,t\onedot} 
\def\etal{\emph{et al}\onedot}
\makeatother

\newcommand{\naive}{na{\"i}ve\xspace} \newcommand{\Naive}{Na{\"i}ve\xspace}
\newcommand{\naively}{na{\"i}vely\xspace} 

\newcommand{\casia}{\mbox{CASIA-WebFace}\xspace}
\newcommand{\cifar}[2][]{\mbox{CIFAR-\num[#1]{#2}}\xspace}

\newcommand{\msceleb}{\mbox{MS-Celeb-\num{1}M}\xspace}
\newcommand{\vggface}{\mbox{VGGFace\num{2}}\xspace}

\newcommand{\resnet}[1]{\mbox{ResNet-\num{#1}}\xspace}
\newcommand{\effnetb}[1]{\mbox{EfficientNet-B\num{#1}}\xspace}

\newcommand{\ccrfpr}[1]{\mbox{CCR@FPR=\SIP{#1}}}
\newcommand{\mixknownratio}{mixup-to-known ratio\xspace}

\newcommand{\MixKnownRatio}{Mixup-to-Known Ratio\xspace}
\newcommand{\mixknownratios}{mixup-to-known ratios\xspace}

\newcommand{\csubrefrange}[2]{\subref{#1}\crefrangeconjunction\subref{#2}}

\DeclareMathOperator*{\argmax}{arg\,max}

\DeclareMathOperator*{\betaD}{Beta}

\newcommand{\kpp}{$K \!+\! 1$\xspace}

\newcommand{\tc}{\text{,}}
\newcommand{\td}{\text{.}}

\newcommand{\Bx}{\bm{x}} 

\newcommand{\Bz}{\bm{z}}

\newcommand{\MC}{\mathcal{C}}
\newcommand{\ME}{\mathcal{E}}

\newcommand{\MR}{\mathbb{R}}

\newcommand{\MT}{\mathcal{T}}

\newcommand{\mytmp}{empty}

\newcommand{\myBelowcaptionskip}{\setlength{\belowcaptionskip}{-2ex}}

\newcommand{\SIP}[1]{\SI{#1}{\percent}}

\newcommand\myMark[1]{\textsuperscript#1}

\makeatletter
\newcommand\thefontsize{The current font size is: \f@size pt}
\makeatother

%% file: settings/glossary.tex
\usepackage[acronym]{glossaries}
\glsdisablehyper

\newcommand*\myglsentry[1]{%
	\protect\ifglsused{#1}{%
		\glsentryshort{#1}%
	}{%
		\glsentrylong{#1}%
	}%
}

\newacronym{roc}{ROC}{Receiver Operating Characteristic}
\newacronym{aucroc}{AUC-ROC}{Area Under the \myglsentry{roc} Curve}
\newacronym{aucoscr}{AUC-OSCR}{Area Under the \myglsentry{oscr} Curve}
\newacronym{cevm}{C-EVM}{Clustering-based \myglsentry{evm}}
\newacronym{cwf}{\mbox{C-WF}}{\casia}
\newacronym{ccr}{CCR}{correct classification rate}
\newacronym{dbscan}{DBSCAN}{Density-Based Spatial Clustering of Applications with Noise}
\newacronym{dir}{DIR}{Detection and Identification Rate}
\newacronym{dnn}{DNN}{Deep Neural Network}
\newacronym{ev}{EV}{extreme vector}
\newacronym{ievm}{iEVM}{incremental \myglsentry{evm}}
\newacronym{evt}{EVT}{extreme value theory}
\newacronym{far}{FAR}{false alarm rate}
\newacronym{fpr}{FPR}{false positive rate}
\newacronym{gmc}{GMC}{Generalized Maximum Coverage}
\newacronym{gan}{GAN}{generative adversarial network}
\newacronym[longplural=known classes]{kc}{KC}{known class}
\newacronym[longplural=known unknown classes]{kuc}{KUC}{known unknown class}
\newacronym{kvr}{KvR}{Known \vs Rest}
\newacronym{lda}{LDA}{linear discriminant analysis}
\newacronym{lfw}{LFW}{Labeled Faces in the Wild}
\newacronym{ml}{ML}{machine learning}
\newacronym{mpl}{MPL}{Multi Pseudo Label}
\newacronym{ncm}{NCM}{Nearest Class Mean}
\newacronym{nno}{NNO}{Nearest Non-Outlier}
\newacronym{oscr}{OSCR}{Open-Set Classification Rate}
\newacronym{ovr}{OvR}{One \vs Rest}
\newacronym{ood}{OOD}{out-of-distribution}
\newacronym{osr}{OSR}{open-set recognition}
\newacronym{owr}{OWR}{open-world recognition}
\newacronym{spl}{SPL}{Single Pseudo Label}
\newacronym{rbf}{RBF}{radial basis function}
\newacronym{tpr}{TPR}{true positive rate}
\newacronym[longplural=unknown unknown classes]{uuc}{UUC}{unknown unknown class}

\newacronym{evm}{EVM}{Extreme Value Machine}

\newacronym{nn}{NN}{Nearest Neighbor}
\newacronym{knn}{$k$-NN}{$k$-\myglsentry{nn}}
\newacronym{osnn}{OSNN}{Open-Set \myglsentry{nn}}
\newacronym{osnno}{OSNN}{Open-Set \myglsentry{nn}}
\newacronym{tnn}{TNN}{Thresholded \myglsentry{nn}}

\newacronym{svm}{SVM}{Support Vector Machine}
\newacronym{wsvm}{W-SVM}{Weibull \myglsentry{svm}}
\newacronym{pisvm}{$P_I$-SVM}{Probability of Inclusion \myglsentry{svm}}

%% file: figures/data/data-commands.tex
\newcommand{\osnnBlCifarAucrocUu}{77.8}
\newcommand{\osnnBlCifarCcrfprUuOne}{28.3}
\newcommand{\osnnBlCifarCcrfprUuTen}{50.4}
\newcommand{\osnnSplGenCifarAucrocUu}{80.0}
\newcommand{\osnnSplGenCifarCcrfprUuOne}{27.1}
\newcommand{\osnnSplGenCifarCcrfprUuTen}{50.3}

\newcommand{\dnnBlCifarAucrocUu}{79.6}
\newcommand{\dnnBlCifarCcrfprUuOne}{23.8}
\newcommand{\dnnBlCifarCcrfprUuTen}{49.9}
\newcommand{\dnnSplGenCifarAucrocUu}{84.1}
\newcommand{\dnnSplGenCifarCcrfprUuOne}{25.5}
\newcommand{\dnnSplGenCifarCcrfprUuTen}{54.8}

\newcommand{\evmBlCifarAucrocUu}{80.0}
\newcommand{\evmBlCifarCcrfprUuOne}{4.4}
\newcommand{\evmBlCifarCcrfprUuTen}{40.1}
\newcommand{\evmSplGenCifarAucrocUu}{82.3}
\newcommand{\evmSplGenCifarCcrfprUuOne}{7.2}
\newcommand{\evmSplGenCifarCcrfprUuTen}{53.4}
\newcommand{\evmKvrGenCifarAucrocUu}{82.6}
\newcommand{\evmKvrGenCifarCcrfprUuOne}{7.2}
\newcommand{\evmKvrGenCifarCcrfprUuTen}{53.4}

\newcommand{\cevmBlCifarAucrocUu}{81.2}
\newcommand{\cevmBlCifarCcrfprUuOne}{6.3}
\newcommand{\cevmBlCifarCcrfprUuTen}{53.2}
\newcommand{\cevmSplGenCifarAucrocUu}{82.9}
\newcommand{\cevmSplGenCifarCcrfprUuOne}{10.5}
\newcommand{\cevmSplGenCifarCcrfprUuTen}{53.8}
\newcommand{\cevmKvrGenCifarAucrocUu}{83.2}
\newcommand{\cevmKvrGenCifarCcrfprUuOne}{10.5}
\newcommand{\cevmKvrGenCifarCcrfprUuTen}{53.8}

\newcommand{\wsvmBlCifarAucrocUu}{80.7}
\newcommand{\wsvmBlCifarCcrfprUuOne}{26.1}
\newcommand{\wsvmBlCifarCcrfprUuTen}{52.6}
\newcommand{\wsvmKvrGenCifarAucrocUu}{82.5}
\newcommand{\wsvmKvrGenCifarCcrfprUuOne}{26.9}
\newcommand{\wsvmKvrGenCifarCcrfprUuTen}{52.9}

\newcommand{\pisvmBlCifarAucrocUu}{82.5}
\newcommand{\pisvmBlCifarCcrfprUuOne}{27.0}
\newcommand{\pisvmBlCifarCcrfprUuTen}{53.2}
\newcommand{\pisvmKvrGenCifarAucrocUu}{83.0}
\newcommand{\pisvmKvrGenCifarCcrfprUuOne}{25.4}
\newcommand{\pisvmKvrGenCifarCcrfprUuTen}{53.2}

%% file: 00paper.tex
\title{\papertitle}

\authors

\maketitle
\ificcvfinal\thispagestyle{empty}\fi

\begin{abstract}
	Handling entirely unknown data is a challenge for any deployed classifier.
	Classification models are typically trained on a static pre-defined dataset and are kept in the dark for the open unassigned feature space.
	As a result, they struggle to deal with out-of-distribution data during inference.
	Addressing this task on the class-level is termed \gls{osr}.
	However, most \gls{osr} methods are inherently limited, as they train closed-set classifiers and only adapt the downstream predictions to \gls{osr}.

	This work presents LORD, a framework to \emph{L}everage \emph{O}pen-set \emph{R}ecognition by exploiting unknown \emph{D}ata.
	LORD explicitly models open space during classifier training and provides a systematic evaluation for such approaches.
	We identify three model-agnostic training strategies that exploit background data and applied them to well-established classifiers.
	Due to LORD's extensive evaluation protocol, we consistently demonstrate improved recognition of unknown data.
	The benchmarks facilitate in-depth analysis across various requirement levels.
	To mitigate dependency on extensive and costly background datasets, we explore mixup as an off-the-shelf data generation technique.
	Our experiments highlight mixup's effectiveness as a substitute for background datasets.
	Lightweight constraints on mixup synthesis further improve \gls{osr} performance.
\end{abstract}
\glsresetall

\placetextbox{0.4875}{0.08}{\copyrightnotice}%

\input{01introduction}

\input{02relatedwork}

\input{03how-to-benchmark}

\input{04how-to-learn}

\input{05how-to-generate}

\input{06discussion}


\clearpage 
{\small
\bibliographystyle{ieee_fullname}
\bibliography{bibliograpy}
}

%% file: 01introduction.tex
\section{Introduction}\label{sec:introduction}
Most classification algorithms are designed for \emph{closed-set} environments, where all classes are known prior to deployment of the classifier.
However, real-world applications may expose classifiers to unseen classes.
Recognizing that an input belongs to an unseen class constitutes the \gls{osr} task.
Scheirer \etal~\cite{scheirer2012openset} distinguish two subtasks:
\begin{enumerate*}
	[label={\arabic*)}]
	\item \emph{Recognizing} samples as known or unknown.
	\item \emph{Classify} knowns to its specific class.
\end{enumerate*}
An example is face recognition of few known subjects among many others~\cite{gunther2017opensetface}.

\begin{figure}[tb]
	\vspace{2pt}
	\renewcommand{\mytmp}{.235}
	\centering
	\tikzsetnextfilename{overview1}%
	\subcaptionbox{Training data (opaque).\label{fig:overview:train}}%
	{\includegraphics[width=\mytmp\textwidth]{figures/overview1}}%
	\hfill%
	\tikzsetnextfilename{overview2}%
	\subcaptionbox{Test data (opaque) with training data (shaded).\label{fig:overview:test}}%
	{\includegraphics[width=\mytmp\textwidth]{figures/overview2}}%
	\myBelowcaptionskip%
	\caption{Overview of data types in \acrlong{osr}.
	The training set in~\subref{fig:overview:train} includes \acrfullpl{kc} and \acrfullpl{kuc}~(\protect\tikzkuc{overview-kuc}).
	The trained classifiers in~\subref{fig:overview:test} model decision boundaries for \acrshortpl{kc} as dashed ellipses.
	KUCs correlate with the training set's KUCs, exhibiting higher identifiability in comparison to the \acrfullpl{uuc}~(\protect\tikzuuc{overview-uuc}), which can exist anywhere in the feature space.}
	\label{fig:overview}%
\end{figure}%

\glsreset{kc}\glsreset{kuc}\glsreset{uuc}%
We distinguish three types of classes~\cite{scheirer2014wsvm} as depicted in \cref{fig:overview}:
\begin{enumerate*}
	[label={\arabic*)}]
	\item \emph{\Glspl{kc}} are uniquely labeled examples detected \emph{and} classified.
	They are essential in both the training and test set.
	\item \emph{\Glspl{kuc}} is background data and comprises samples not necessarily grouped into meaningful categories, but assumed to be unrelated to the \glspl{kc}.
	They are part of the training and test set.
	We differentiate \emph{genuine} \glspl{kuc}, a dataset subset, from \emph{synthesized} \glspl{kuc}.
	\item \emph{\Glspl{uuc}} are unseen during training.
	They are only part of the test set.
\end{enumerate*}

In recent years, \gls{osr} has gained increasing attention~\cite{geng2020osrsurvey,salehi2021oodsurvey,yang2021oodsurvey}.
Most approaches adopt \emph{closed-set training with open-set inference}, as exemplified in various methods, including distance-based classifiers~\cite{bendale2015openworld,junior2017osnn,mensink2013ncm,ristin2014incm}, \glspl{svm}~\cite{bartlett2008classification,grandvalet2008svm,jain2014pisvm,scheirer2014wsvm}, \glspl{evm}~\cite{henrydoss2020cevm,koch2022ievm,rudd2017evm,vignotto2018evm_gpd_gev}, losses and calibrations for \glspl{dnn}~\cite{bendale2016opensetnn,miller2021cacloss,shu2017doc}, uncertainty quantification~\cite{blundell2015bnn,lorch2020jpeg,lorch2021gps}, novelty detection~\cite{boult2022evtwow}, and auto-encoder architectures~\cite{oza2019c2ae,perera2020generative,yoshihashi2019crosr}.
For inference, a threshold is introduced to either reject or classify a test sample based on a confidence value~\cite{tax2008growing}.

Let $\MT_K \!=\! \{(\Bx_i, y_i)\}_{i=1}^{T}$ be a labeled training dataset with features $\Bx \!\in\! \MR^D$ and class labels $y \!\in\! \MC_K$, where $\MC_{K}$ denotes the set of \glspl{kc}.
Given $\MT_K$, the goal is to infer a likelihood function $f(\Bx; \MT_K) \!=\! \Bz$ with $\Bz \!\in\! \MR^{|\MC_{K}|}$ that maps features to a confidence $z_y \!\in\! \MR$ for each class in $\MC_{K}$.
For instance, $\Bz$ could describe normalized posterior probabilities or distances.
Function $f$ is used to decide for class $\hat{y}$ with the highest confidence: \mbox{$\hat{y} \!=\! \argmax_{\hat{y} \in \MC_{K}} f(\Bx; \MT_K)_{\hat{y}}$}.
A reject option converts this prediction into \gls{osr} according to:
\begin{equation}
	\label{eq:openset1}
	g(\Bx, \delta; \MT_K) =
	\begin{cases}
		\hat{y} & \text{if } \max_{\hat{y} \in \MC_{K}} f(\Bx; \MT_K)_{\hat{y}} > \delta \enspace \tc\\
		u & \text{otherwise} \enspace \tc
	\end{cases}
\end{equation}
where $\delta$ denotes the decision threshold.
However, training exclusively on a closed set can benefit the classification task but inherently limits the \gls{osr} capabilities.

To the best of our knowledge, only a few works consider background samples~\cite{dhamija2018oscr,gunther2017opensetface,perera2019deep}.
While genuine \glspl{kuc} undoubtedly simplify \gls{osr}, they may not be available.
Recent approaches show that synthesizing \gls{ood} data with \glspl{gan} can suffice for training \glspl{dnn}~\cite{ge2017gopenmax,neal2018counterfactual,schlachter2019intraclass_split}.
However, \gls{gan} training can be unstable~\cite{kong2021opengan} and is designed for in-distribution data~\cite{goodfellow2014gan}, necessitating additional methods to handle outliers~\cite{ge2017gopenmax}.
The mentioned works offer valuable insights into end-to-end \gls{osr} with \glspl{kuc}.
However, most are trained in a \kpp fashion, where the background is modeled in a single class.
This approach is limited by the background class's complexity~\cite{dhamija2018oscr}.
An exception is the work by G{\"u}nther \etal~\cite{gunther2017opensetface} that studies \gls{osr} via \acrlong{lda} and the \gls{evm}.
We extend their work by exploiting \glspl{kuc} in diverse ways and examine whether synthetic data is a suitable proxy for genuine \glspl{kuc}.

Our framework LORD provides guidance for researchers to explicitly model open space during classifier training.
To verify the efficacy of this modeled space on unknown recognition, a benchmark protocol is introduced.
The protocol requires, besides \glspl{kc}, access to genuine \glspl{kuc} and \glspl{uuc}.
LORD distinguishes between two assessments:
\begin{enumerate*}
	[label={\arabic*)}]
	\item biased assessment, involving training with genuine \glspl{kuc} and testing with genuine \glspl{kuc} and \glspl{uuc}.
	\item Unbiased assessment excludes \glspl{kuc} during testing.
\end{enumerate*}
This identifies optimal learning conditions for different models and application-specific requirements.
In the context of face recognition, a biased application could be a system at an airport likely to encounter genuine \glspl{kuc}, while an unbiased application might be the authentication on a smartphone.

LORD encompasses a comprehensive range of \gls{osr} metrics that facilitate in-depth analysis across various requirement levels.
These levels may place a particular emphasis on secure open-set performance or prioritize strong closed-set performance, catering a more user-friendly behavior.
In particular, this extensive evaluation highlights a fundamental trade-off between open- and closed-set performance across models.

To model open space, LORD deploys three model-agnostic training strategies on six classifiers spanning four distinct categories.
These strategies improve biased \gls{osr} performance by up to \SIP{30}, while unbiased performance generally demonstrates a more modest improvement.

Admitting the scarcity and cost of domain-specific background data in real-world applications, LORD offers a solution in the form of mixup as an off-the-shelf data generation technique.
Mixup samples are convex combinations of distinct \glspl{kc} and act as substitutes for genuine \glspl{kuc}.
While \naive mixups constitute the occupation problem, LORD proposes effective constraints to refine the generation process.
Our experiments confirm mixup as an excellent substitute for genuine \glspl{kuc}.
This approach proves beneficial in assessing models before investing in resource-intensive collection of background data.

To summarize, LORD provides a systematic framework to assess the efficacy of background data exploitation to enhance classifiers within particular application contexts.
When background data is not available, mixup poses an effective solution.
If this proves beneficial, researchers might find it valuable to gather additional background data.

This work is organized as follows:
\Cref{sec:relatedwork} reviews related works.
\Cref{sec:3how-to-benchmark} introduces our \gls{osr} benchmarking protocol.
\Cref{sec:4-how-to-learn} outlines the training strategies and their use for $6$ \gls{osr} models.
\Cref{sec:5-generate} explores mixup and the constrained generation.
\Cref{sec:discussion} summarizes the main findings and concludes this work.

%% file: 02relatedwork.tex
\section{Related work}\label{sec:relatedwork}
We explore incorporating \glspl{kuc} during training and reformulate \cref{eq:openset1} accordingly.
Let \mbox{$\MT = \{(\Bx_i, y_i)\}_{i=1}^{T}$} be a \emph{partially} labeled training dataset with class labels \mbox{$y  \in  \MC$}, where $\MC  =  \MC_{K}  \cup  u$ and $u$ indicates a missing label.
Set $\MT$ contains samples of \glspl{kc} and \glspl{kuc}, leading to a change in function $g$ where $\MT_K$ is replaced by $\MT$, \mbox{$h(\Bx, \delta; \MT) = g(\Bx, \delta; \MT)$}.
The class decision remains unchanged, but the inclusion of \glspl{kuc} impacts the likelihood function $f$, categorizing different approaches in \gls{osr}.

\subsection{Semi-supervised model training}
\label{subsec:semi-supervised-model-training}

\Glspl{kuc} samples can also be considered as a special instance of unlabeled data.
Exploiting both labeled and unlabeled data is a common focus in semi-supervised learning.
Notable techniques in this field comprise manifold regularization~\cite{melacci2011laplaciansvm} and the generation of pseudo-labels~\cite{arazo2020pseudolabel,hu2021simple}, integrating unlabeled data into a supervised learning regime.
Using unlabeled samples is considerably cheaper than labeled samples~\cite{wang2020focalmix}.
Jointly exploiting labeled and unlabeled data can substantially enhance the underlying models compared to purely supervised learning~\cite{zhang2021flexmatch}.

However, unlike \gls{osr}, semi-supervised learning tackles a complementary scenario: predictions cover only \glspl{kc}.
The approach assumes that unlabeled instances belong to one or more of the \glspl{kc}, without considering the open space.
We follow concepts similar to semi-supervised learning and incorporate unlabeled \glspl{kuc} to improve \gls{osr} learning.

\subsection{Open-set training with open-set inference}
\label{subsec:open-set-train-open-set-inference}

The vast body of previous works use unlabeled \glspl{kuc} in a \kpp fashion, \ie, $K$ \glspl{kc} and one unknown class~\cite{ge2017gopenmax,neal2018counterfactual,schlachter2019intraclass_split}.
In function $h$, $\MC_K$ is substituted by $\MC$, treating all samples with the label $u$ as one large class.
This allows the model to predict the background class without a threshold $\delta$.
However, we show that modeling unknowns in a single class can be suboptimal and is surpassed by alternative strategies.

Others use external datasets to define special loss functions to calibrate filters~\cite{perera2019deep} or probabilities~\cite{dhamija2018oscr}, particularly tailored to train \glspl{dnn}.
Günther \etal~\cite{gunther2017opensetface} take an initial step towards employing genuine \glspl{kuc} with \glspl{evm}.
They do not cluster \glspl{kuc} but consider them to learn probabilistic models of \glspl{kc}.
This can improve the detection of \glspl{uuc} compared to simple baselines like distance-based detection.
These findings indicate the potential of leveraging \glspl{kuc} to enhance the predictive power of \gls{osr} models.

In this paper, we generalize model-specific approaches into three different model-agnostic training strategies that exploit \glspl{kuc}.
Specifically, we extend multiple \gls{osr} models using these training strategies and determine the optimal strategy based on the underlying model and task.

\subsection{Generating out-of-distribution samples}
\label{subsec:generation-of-out-of-distribution-samples}

Recent works addressed the question of whether \glspl{kuc} could be synthesized when external data is unavailable.
For instance, Du \etal~\cite{du2021vos} model \glspl{kuc} by estimating Gaussian distributions per \gls{kc} and sampling background data near class boundaries.
However, the Gaussian distribution assumption is simplistic, resulting in uniform restrictions of the decision boundary.
Zhou \etal~\cite{zhou2021proser} regularizes \gls{dnn} training using manifold mixup~\cite{verma2019manifoldmixup} to learn placeholders.

In-distribution data is synthesized through generative models~\cite{davari2018gmmsynth,goodfellow2014gan,perez2017effectiveness,shrivastava2017learning} or augmentation~\cite{cubuk2019autoaugment,lim2019fastautoaugment,perera2021geoxtreme,verma2019manifoldmixup,yun2019cutmix,zhang2018mixup}.
Conversely, \gls{ood} data generation is more difficult.
Ge \etal~\cite{ge2017gopenmax} propose a complex pipeline:
\begin{enumerate*}
	[label={\arabic*)}]
	\item Train a \gls{dnn} and conditional \gls{gan} using \glspl{kc}.
	\item Generate \gls{gan} samples and classify them with the \gls{dnn}.
	\item Store misclassified samples as \glspl{kuc}.
	\item Train another \gls{dnn} in a \kpp way with \glspl{kuc} as background class.
\end{enumerate*}
This generation is guided by the assumption that regions where both models disagree do not belong to any \gls{kc}.
Unfortunately, achieving convergence of one \gls{gan} is challenging, leading to the training of several \glspl{gan} on different class subsets.
Also, control over the generated random samples is very limited, and the required number of \glspl{kuc} remains an open question.
Neal \etal~\cite{neal2018counterfactual} train an encoder-decoder \gls{gan} and a classifier.
To generate \glspl{kuc}, they alter KC samples in the latent space of the \gls{gan} to make the \gls{dnn} uncertain about the prediction.
The \gls{dnn} is finetuned in a \kpp manner.
Also here, there is a mutual dependence between the \gls{gan} and \gls{dnn} and the generation process lacks control.
Generating images is challenging due to their high dimensionality and the need to match the data context.
Moreover, \glspl{dnn} are vulnerable to small deviations as introduced by adversarial attacks~\cite{akhtar2018adversarialsurvey,eykholt2018physicattack}.

Kong and Ramanan~\cite{kong2021opengan} avoid mutual dependencies using DNN penultimate layer features to train a \gls{gan}.
The \gls{gan}'s discriminator detects unknowns, while the \gls{dnn} classifies knowns.
We adopt a similar two-stage approach, but unlike \glspl{gan} operating in image space, we operate solely in the feature space.
Specifically, we use manifold mixup~\cite{verma2019manifoldmixup} to generate \glspl{kuc}, a lightweight alternative to complex GANs.
However, manifold mixup may generate unwanted \glspl{kuc} that overlap with a \gls{kc}, referred to as \textit{occupation problem}.
To mitigate this, we introduce an effective filter to retain only meaningful \glspl{kuc}.

%% file: 03how-to-benchmark.tex
\section{How to benchmark with known unknowns?}\label{sec:3how-to-benchmark}
Common \gls{osr} benchmarks~\cite{dhamija2018oscr,neal2018counterfactual,scheirer2012openset} ignore \glspl{kuc} for training and testing.
In contrast, we extend the work of G{\"u}nther \etal~\cite{gunther2017opensetface} to explore the impact of \glspl{kuc} on model training.
This serves as a reference under conditions when dataset-related background data is available.


\subsection{Defining protocols with known unknowns}\label{subsec:protocol}
For our experiments, we define evaluation protocols using well-established datasets with varying characteristics.

\paragraph{\cifar[math-rm=\mathbf]{100}~\textmd{\cite{krizhevsky09cifar}}.}

This image dataset consists of $20$ superclasses divided into $100$ equally balanced subclasses.
We split the superclasses into \num{12}~\glspl{kc}, \num{4}~\glspl{kuc}, and \num{4}~\glspl{uuc}.
Adopting the subclass labels, this results in \num{60}~\glspl{kc}, \num{20}~\glspl{kuc}, and \num{20}~\glspl{uuc}.
The training set contains \num{500}~images per class, leading to a \gls{kuc} to \gls{kc} sample ratio of \num{0.33}.
For feature extraction, we use \effnetb{4}~\cite{tan2019efficientnet}, pre-trained on ImageNet~\cite{deng2009imagenet} and finetuned on the $60$~\glspl{kc}.
The embedding dimension is~\num{1792}.
Unless otherwise stated, we repeat all experiments $3$~times with varying superclass-splits and report averaged results.
While a pre-trained ImageNet feature extractor already saw all \cifar{100} classes, this setup remains open-set for \num{3} reasons:
\begin{enumerate*}
	[label={\arabic*)}]
	\item Image resolutions significantly differ.
	\item Finetuning is subject to catastrophic forgetting~\cite{mccloskey1989catastrophic,rebuffi2017icarl}.
	\item We use the features to train a downstream classifier, which is never exposed to ImageNet.
\end{enumerate*}

\glsreset{lfw}  

\paragraph{\Gls{lfw}~\textmd{\cite{huang2014lfw,huang2007lfw}}.}
For this face recognition dataset, we follow the protocol of G{\"u}nther \etal~\cite{gunther2017opensetface} and evaluate the $o_3$~probe set.
The training set consists of $3$~samples for each of the $610$~\glspl{kc} and $1$~sample for each of the \num{1070}~\glspl{kuc} which is a ratio of \num{0.58}.
We extract \num{128}-dimensional features using \resnet{50}~\cite{he2016resnet}, pre-trained on \msceleb~\cite{guo2016msceleb}, and finetuned on \vggface~\cite{cao2018vggface2}.

\glsreset{cwf}  

\paragraph{\Gls{cwf}~\textmd{\cite{yi2014casia}}.}
This is a face recognition dataset with imbalanced distributions.
We split it into \num{6345}~\glspl{kc}, \num{2115}~\glspl{kuc}, and \num{2115}~\glspl{uuc}.
Unless otherwise stated, all experiments are repeated $3$ times with varying \gls{kc} and \gls{kuc} splits and averaged results are reported.
On average, the least represented \gls{kc} has \num{13}~samples and the largest \num{627}~samples.
The KUC to KC sample ratio is \num{0.33}.
Feature extraction uses the same model as for \gls{lfw} but with a feature dimension of \num{2048}.
We also employ a reduced version (Tiny \gls{cwf}) to address classifier scaling issues with the full dataset.
Tiny \gls{cwf} retains \SIP{20} of \glspl{kc}, \glspl{kuc}, and \glspl{uuc}.
The number of samples per class is reduced by \SIP{50}.

\subsection{Evaluating open-set recognition models}\label{subsec:evaluatingopensetrecognition-models}
First, we assess the ability to distinguish knowns from unknowns in a binary problem.
This is supported by a \gls{roc} depicting the \gls{tpr} \vs the \gls{fpr}.
To obtain a performance measure independent of the threshold $\delta$, we use the \gls{aucroc}.

Second, we employ the \gls{oscr}~\cite{dhamija2018oscr}, which measures the \gls{ccr} and \gls{fpr}.
Let $g(\Bx,\delta;\MT)$ be a decision function as formalized in \cref{eq:openset1}.
The test set comprises three types of classes, denoted as $\ME  =  \ME_{K} \cup \ME_u$ with $\ME_u  =  \ME_{KU} \cup \ME_{UU}$.
Then these measures are expressed as follows:
\begin{align}
	\!\!\text{CCR}(\delta) \!&=\! |\ME_{K}|^{-1}|\{\Bx | (\Bx, y) \in \ME_{K} \land g(\Bx,\delta;\MT) \!=\! y\}| \, \tc\\
	\!\!\text{FPR}(\delta) \!&=\! |\ME_{u}|^{-1}|\{\Bx | \Bx \in \ME_{u} \land g(\Bx,\delta;\MT) \neq u\}| \, \td \label{eq:fpr}
\end{align}
For most open-set applications, \eg face recognition, a high \gls{ccr} at low \gls{fpr} is preferable~\cite{gunther2017opensetface}.
Therefore, we report the \glspl{ccr} in the open-set relevant \gls{fpr} range of \SIrange{0}{10}{\percent}.

\subsection{Biased \vs unbiased evaluation}\label{subsec:biased-vs-unbiased-evaluation}
The biased protocol in \cref{subsec:protocol} with genuine \glspl{kuc} in the training set simplifies the detection of \glspl{kuc} in the test set, resulting in significant improvements in \gls{osr} measures.

To ensure unbiased evaluation, we compute the metrics solely on a test subset comprising \glspl{kc} and \glspl{uuc}, excluding \glspl{kuc}.
Thus, $\ME_{u}$ in \cref{eq:fpr} is substituted by $\ME_{UU}$.
\emph{Biased} evaluation considers metrics for both \glspl{kuc} and \glspl{uuc}, while \emph{unbiased} evaluation focuses solely on \glspl{uuc}.
The biased assessment tends to paint an overly optimistic picture for applications that primarily deal with \glspl{uuc}.

%% file: 04how-to-learn.tex
\section{How to learn models with known unknowns?}\label{sec:4-how-to-learn}
This section outlines three open-set training strategies and their deployment to six classifiers of four categories.


\subsection{Open-set training strategies}\label{subsec:strategies}
\begin{figure}[tb]
	\centering
	\renewcommand{\mytmp}{0.119}%
	\subcaptionbox{EVM\label{fig:strategies:ignore}}[\mytmp\textwidth]%
	{\includegraphics[width=\linewidth]{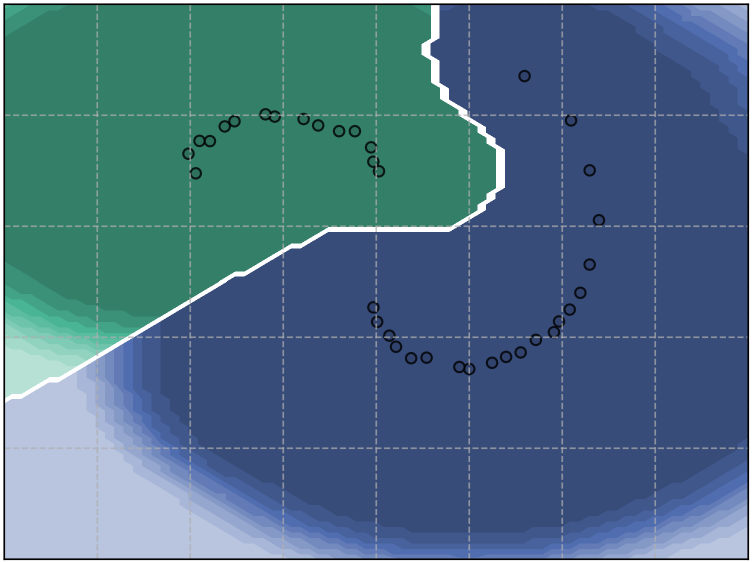}}\hfill%
	\subcaptionbox{+ \acrshort{spl}\label{fig:strategies:single}}[\mytmp\textwidth]%
	{\includegraphics[width=\linewidth]{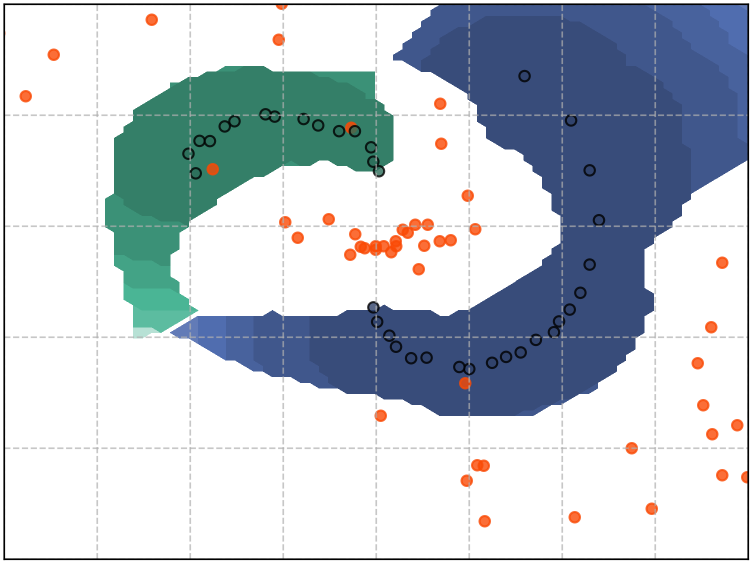}}\hfill%
	\subcaptionbox{+ \acrshort{mpl}\label{fig:strategies:multi}}[\mytmp\textwidth]%
	{\includegraphics[width=\linewidth]{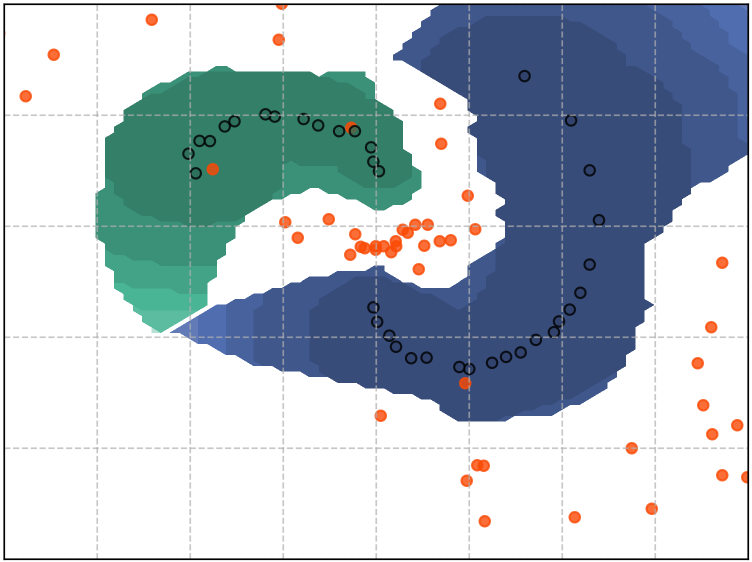}}\hfill%
	\subcaptionbox{+ \acrshort{kvr}\label{fig:strategies:negative}}[\mytmp\textwidth]%
	{\includegraphics[width=\linewidth]{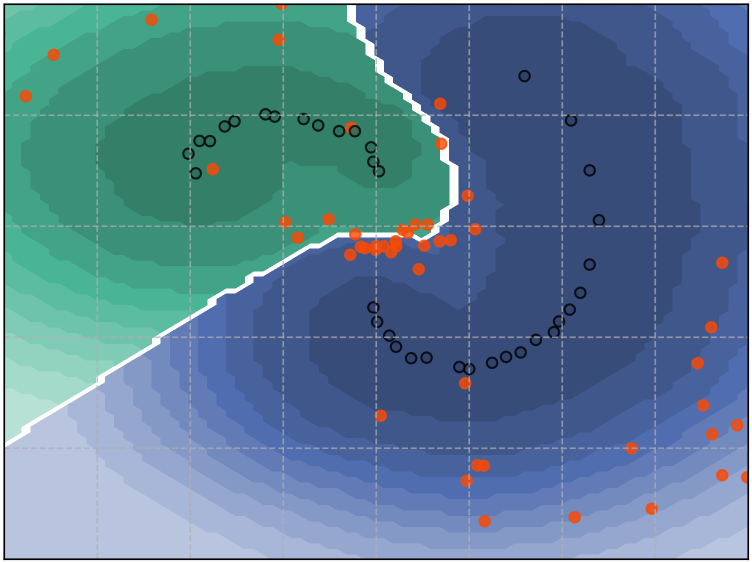}}%
	\myBelowcaptionskip%
	\caption{EVM decision boundaries of training strategies exploiting \acrfullpl{kuc}. This is a toy dataset with dark-edged points for $2$ \acrfullpl{kc} and bright orange points for \acrshortpl{kuc}. Colored areas belong to the related class with confidence visualized via opacity, \ie, white is \emph{zero} confidence.}\label{fig:strategies}%
\end{figure}

Incorporating \glspl{kuc} into the training process of classifiers can be achieved using various strategies.
\Cref{fig:strategies} illustrates examples for these approaches, including the \gls{evm}.

\glsreset{spl}  

\paragraph{\Gls{spl}.}
Samples without labels are assigned the \emph{pseudo} label $u$ and treated as a single class.
This method aligns with the \kpp training strategy in previous works~\cite{ge2017gopenmax,neal2018counterfactual,schlachter2019intraclass_split} with function $h$, but with $\MC_K$ substituted by $\MC$.
This also implies $f(\Bx; \MT)  =  \Bz$ with \mbox{$\Bz  \in  \MR^{|\MC|}$}.

The model can now predict label $u$ directly independent of the threshold $\delta$.
This results in steep decision boundaries as demonstrated in \cref{fig:strategies:single}.
While this approach is universally applicable, the complex distribution of the large background class presents a challenge.
Depending on the underlying \gls{osr} model, finding suitable representations for the background class formed by the pseudo label is difficult.

\begin{figure*}[tb]
	\centering
	\includegraphicstikz{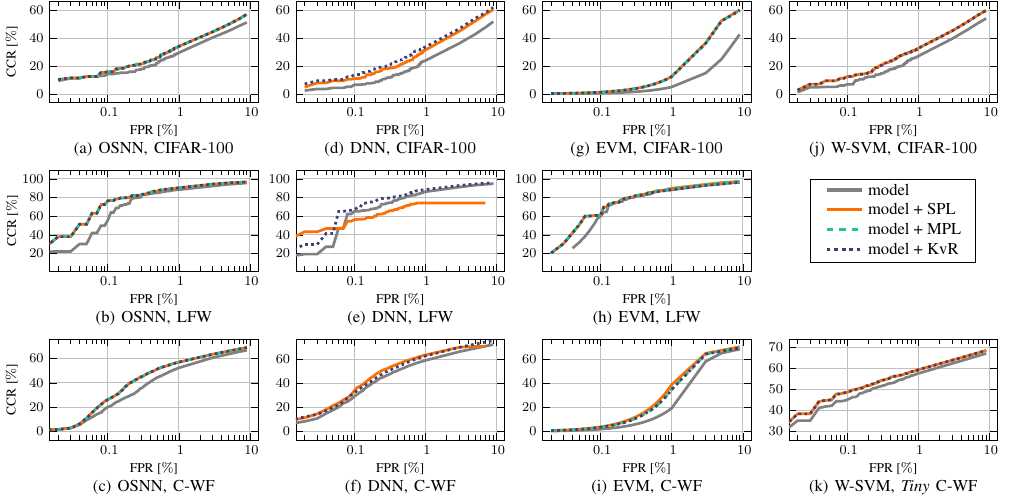}{.98\linewidth}%
	\myBelowcaptionskip%
	\caption{Results of the biased evaluation of $4$~models (column-wise) exploiting genuine KUCs with the training strategies and the baseline on $3$ datasets (row-wise).
	The models comprise OSNN in~\csubrefrange{subfig:osnn-genuine-kuc-group-cifar}{subfig:osnn-genuine-kuc-group-casia}, DNN in~\csubrefrange{subfig:dnn-genuine-kuc-group-cifar}{subfig:dnn-genuine-kuc-group-casia}, EVM in~\csubrefrange{subfig:evm-genuine-kuc-group-cifar}{subfig:evm-genuine-kuc-group-casia}, and \acrshort{wsvm} in~\subref{subfig:wsvm-genuine-kuc-group-cifar} and \subref{subfig:wsvm-genuine-kuc-group-casia}.
	Results in~\subref{subfig:wsvm-genuine-kuc-group-casia} are for \emph{Tiny} \acrshort{cwf}.
	Shown is the biased \acrfull{oscr} in the open-set relevant FPR range up to \SIP{10}.}
	\label{fig:genuine-kuc-super}%
\end{figure*}%

\glsreset{mpl}  

\paragraph{\Gls{mpl}.}
This strategy assigns an individual pseudo label to each \gls{kuc} sample.
The number of classes in $\MC$ increases by $|\MT_{u}|$, the amount of \gls{kuc} samples in $\MT$.
The remaining strategy follows the \gls{spl}, but now each former \gls{kuc} contains only one sample.
If one of the newly declared classes is predicted, it must be mapped to label $u$.

Similar to \gls{spl}, this approach enables direct predictions of unknown labels, resulting in sharp decision boundaries, \cf \cref{fig:strategies:multi}.
We hypothesize that multiple pseudo labels stabilize \gls{osr} learning, especially when each \gls{kuc} sample belongs to a distinct category with high inter-category distances in feature space.
A notable drawback is the significant increase in classes, making it impracticable for \gls{ovr} models.
Additionally, \gls{mpl} may induce label noise, leading to inconsistent decision boundaries.

\glsreset{kvr}  

\paragraph{\Gls{kvr}.}
This mode is akin to the common \gls{ovr} multiclass strategy, often used for binary classifiers like \glspl{svm} to handle multiclass problems.
In \gls{kvr}, each \gls{kuc} never acts as a positive class, avoiding the need for its own model representation.
Instead, it serves as negative in the rest-class of other binary models.
This approach resolves the issue of complex background class representation.
Other classes can still adjust their decision boundaries to the negative samples, observed in \cref{fig:strategies:negative} where \glspl{kc}' boundaries are tightened by \glspl{kuc}.
Implementing \gls{kvr} can vary, but we intend for the background data not to be explicitly modeled as a pseudo-class.
It serves as a tool to determine better decision boundaries for \glspl{kc} concerning the open space.
Experiments in \cref{subsec:doknownunknownsimproveosr} show that this approach enhances unknown recognition.

\glsreset{osnn}
\glsreset{dnn}
\glsreset{evm}
\glsreset{svm}
\glsreset{pisvm}
\glsreset{wsvm}

\subsection{Deployed strategies for open-set models}\label{subsec:applying-strategies}
In this section, we briefly outline how to deploy the strategies to \num{6} \gls{osr} models representing \num{4} distinct categories.
More details are provided in the supplements.

The \gls{osnn}~\cite{junior2017osnn} exploits the ratio between the two nearest samples from distinct classes as confidence value.
\gls{spl} includes all \glspl{kuc} in distance ratio computation and class prediction.
\Gls{mpl} treats each \gls{kuc} as a separate class, impacting the reject option only.
For \gls{kvr}, only \glspl{kc} are used for label prediction, while both \glspl{kc} and \glspl{kuc} determine distance ratios.
The strategies primarily vary in the confidence computation, leading to nearly identical results.

Given the vast possibilities of training \glspl{dnn} with \glspl{kuc} by tailoring losses, we opt for the classical cross-entropy loss.
For \gls{spl}, we expand the number of output units by one class.
\Gls{mpl} does not scale to large \gls{kuc} sets.
Entropic open-set loss~\cite{dhamija2018oscr} is adopted~for~\gls{kvr}.

The \gls{evm}~\cite{rudd2017evm}, C-EVM~\cite{henrydoss2020cevm}, and both \gls{svm} variants, \gls{wsvm}~\cite{scheirer2014wsvm} and \gls{pisvm}~\cite{jain2014pisvm}, implement an \gls{ovr} scheme.
For each \emph{one}-class the other \emph{rest}-classes serve as counterpart.
For these models \gls{spl} and \gls{mpl} differ only in the number of pseudo-classes.
The \glspl{svm} omit \gls{mpl} due to its limited scalability to large number of classes and their requirement of at least three samples per class.
The \gls{kvr} strategy can be deployed by not representing the \glspl{kuc} as a positive one-class.
Instead, they are always considered part of the rest-class.

\subsection{Performance of training strategies}\label{subsec:doknownunknownsimproveosr}
We benchmark the training strategies using the protocols~in~\mbox{\cref{subsec:protocol}}.
Hyperparameters are determined through a grid search based on $5$-fold stratified cross-validation.
For LFW, standard $3$-fold cross-validation is applied.
The grid search is conducted for the baseline models and the parameters are reused for \gls{spl}, \gls{mpl}, and \gls{kvr}.
This approach may favour the baseline, where no \glspl{kuc} are used.

\begin{figure*}[tb]
	\centering
	\includegraphicstikz{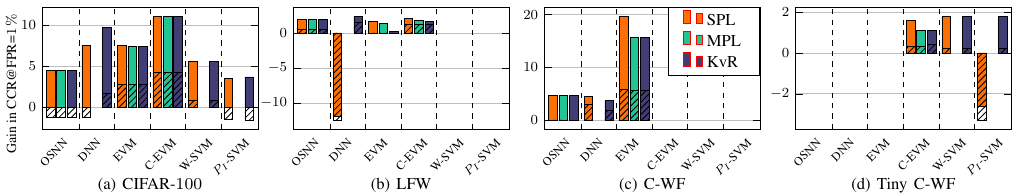}{.99\linewidth}%
	\myBelowcaptionskip%
	\caption{Biased \vs unbiased OSR performance across all models, strategies, and datasets.
	Given is the gain in the \ccrfpr{1} of the strategies using genuine KUCs over the respective baseline model trained without KUCs.
	The colored bars show the gain in the biased evaluation, \ie the performance over KCs, KUCs, and UUCs.
	Shaded areas show the gain in the unbiased evaluation with excluded KUCs.}
	\label{fig:biased-vs-unbiased-bar}%
\end{figure*}%

\paragraph{Biased evaluation.}
In \cref{fig:genuine-kuc-super}, we report average \gls{oscr} measures for the three datasets.
Results for the C-EVM and \acrshort{pisvm} are reported in the supplement.
\Gls{kuc} exploitation improves \gls{osr} performance compared to baselines without \glspl{kuc}.
The exact behavior is model-dependent.

For \gls{osnn}, \cf \crefrange{subfig:osnn-genuine-kuc-group-cifar}{subfig:osnn-genuine-kuc-group-casia}, we observe consistent improvements by \gls{spl}, \gls{mpl}, and \gls{kvr}.
The highest gain of about \SIP{21} is observed in \cref{subfig:osnn-genuine-kuc-group-lfw} for FPRs below \SIP{0.1}.
This result is crucial for safety-relevant applications that require low \glspl{fpr}.
For \gls{cwf}, we see a gain in the \gls{fpr} range of \SIrange{0.1}{1}{\percent}, and for \cifar{100} from \SIP{0.1} upwards.
The exploitation of \glspl{kuc} does not negatively impact the \gls{ccr}.

For \gls{dnn}, \cf \crefrange{subfig:dnn-genuine-kuc-group-cifar}{subfig:dnn-genuine-kuc-group-casia}, \gls{kvr} consistently outperforms the baseline.
\Gls{spl} performs similarly to \gls{kvr} for \cifar{100} and \gls{cwf} but is outperformed by \gls{kvr} and the baseline on \gls{lfw}, except for FPRs below \SIP{0.1}.
We explain this behaviour by the few samples per \glspl{kc} and \gls{kuc} in case of \gls{lfw}, causing two problems:
\begin{enumerate*}
	[label={\arabic*)}]
	\item Each KUC is represented by one sample, making it challenging to model a pseudo-class.
	\item The pseudo-class is large compared to the KCs.
\end{enumerate*}
After examining \gls{spl}, we found that the \gls{dnn} mainly predicts unknowns, a common issue of \glspl{dnn} with unbalanced classes~\cite{buda2018cnnimbalance}.
This is not the case for \gls{cwf} and \cifar{100}, where all classes have sufficient samples.

For the \gls{evm} in \crefrange{subfig:evm-genuine-kuc-group-cifar}{subfig:evm-genuine-kuc-group-casia}, all training strategies outperform the baseline.
For \cifar{100}, we observe a gain of \SIP{17.6} in the \ccrfpr{10}, for \gls{lfw} a gain of \SIP{30.6} at an \gls{fpr} of \SIP{0.05}, and for \gls{cwf} \SIP{20} at an \gls{fpr} of \SIP{1}.

The \gls{wsvm} results for \cifar{100} and Tiny \gls{cwf} are in \cref{subfig:wsvm-genuine-kuc-group-cifar,subfig:wsvm-genuine-kuc-group-casia}.
As indicated in \cref{subsec:applying-strategies}, we do not apply the \gls{mpl} strategy and also do not evaluate LFW due to the model's limitations.
By exploiting KUCs, \gls{osr} performance in the selected open-set range is improved by up to \SIP{6} and \SIP{4} for \cifar{100} and Tiny \gls{cwf}, respectively.

\paragraph{Unbiased evaluation.}

The question remains whether \glspl{kuc} also improve the detection of \glspl{uuc}.
In \cref{fig:biased-vs-unbiased-bar}, we compare the biased and unbiased evaluation using the \ccrfpr{1} gain in relation to the baseline.
The performances expressed by the unbiased evaluation fall below the biased measures.
This indicates that \glspl{uuc} are more difficult to detect and the biased evaluation draws a too optimistic picture.
We found that this discrepancy is model-dependent.
The OSNN exhibits a loss for \cifar{100} in the unbiased measure and only marginal gains for LFW and \gls{cwf}.
The DNN with KvR outperforms the baseline in both evaluations.
This supports the finding that the entropic open-set loss is more suitable for OSR than the simple \kpp (SPL) strategy \cite{dhamija2018oscr}.
The \glspl{evm} improves over the baselines in both biased and unbiased cases, except in for LFW.
Unlike the \gls{osnn}, the \glspl{evm} take advantage of a tail of \gls{kc} and \gls{kuc} samples to refine the decision boundary of \glspl{kc}.
In contrast to the \gls{wsvm}, the \gls{pisvm} does not always benefit from \glspl{kuc} in the unbiased case.

%% file: 05how-to-generate.tex
\section{How to generate known unknowns?}\label{sec:5-generate}
We showed that genuine \glspl{kuc} can improve \gls{osr}.
However, in certain domains, genuine \gls{kuc} data may be unavailable.
This raises the question of how to synthesize them.
In contrast to \glspl{gan} that are not easily usable as standalone \gls{ood} generator for arbitrary \gls{osr} models, \cf \cref{subsec:open-set-train-open-set-inference}, our proposed framework answers this question using the tools at hand, namely the data and the feature extractor.


\subsection{Mixup -- known unknowns straight off the shelf}\label{subsec:ismixupsuitable}
Mixup~\cite{zhang2018mixup} is a lightweight augmentation technique for \glspl{dnn} that generates new data by convex combinations of samples.
Jian \etal~\cite{jiang2023openmix+} show that standard mixup increases open-space risk by smoothing decision boundaries between classes.
However, we further constrain mixup to use samples from distinct classes and assign to the mixed sample $\tilde{\Bx}$ the unknown label $u$, leading to steep decision boundaries.
Here, $(\Bx_i,y_i)$ and $(\Bx_j, y_j)$ are randomly drawn from $\MT_{K}$:
\begin{equation}
	\tilde{\Bx} = \lambda \Bx_i + (1 - \lambda) \Bx_j \enspace \text{, with} \enspace y_i \neq y_j \enspace \text{and} \enspace \tilde{y} = u \enspace \td \label{eq:mixup}
\end{equation}

We use \emph{manifold mixup}~\cite{verma2019manifoldmixup}, as mixing in feature space offers several advantages:
\begin{enumerate*}
	[label={\arabic*)}]
	\item It is more efficient, as the combinations no longer need to be computed in the high-dimensional image space.
	\item It is independent of peculiarities of the feature extractor, eliminating the need to pass the mixed image through that extractor.
\end{enumerate*}

We use \cifar{100} and replace genuine \glspl{kuc} with mixups of \glspl{kc}.
Mixing factor~$\lambda$ in \cref{eq:mixup} is set to \mbox{$\lambda \sim \betaD(2, 2)$} with \mbox{$\lambda \in [0.4, 0.6]$}, ensuring that most mixups lie between two classes.
While training with synthetic and testing with genuine \glspl{kuc} is not biased, we present the metrics calculated on the test set with excluded \glspl{kuc} for a fair comparison with models using genuine \glspl{kuc}, and we continue to refer to this as \emph{unbiased}.
We also assess \gls{osr} performance by the number of generated mixup samples, represented by the \emph{\mixknownratio}, indicating how many mixups are generated per one known sample.

\input{tables/evm-toy-3class-vertical.tex}


\subsection{Augmenting models by manifold mixup}

We exemplary investigate manifold mixup for the \gls{evm} in \cref{tab:evm-toy-3class}.
With more mixups, the gaps between classes are filled and the open space is extended.
The strategies have varying degrees of invasiveness, with \gls{spl} favoring open space the most, followed by \gls{mpl} and then \gls{kvr}.
\gls{kvr} requires finding a suitable threshold to form open space.

\begin{figure}[tb]
	\centering
	\includegraphicstikz{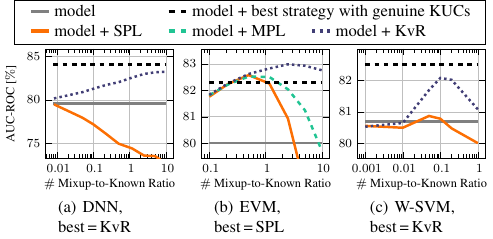}{\linewidth}%
	\myBelowcaptionskip%
	\caption{Unbiased \cifar{100} results of models exploiting mixups.
	Shown are the $3$ strategies with the \acrshort{aucroc} over the number of mixup samples.
	The baseline model without KUCs~(\tikzref{cifar100-mfmix-supergroup-bl}) and the best strategy of each model trained with genuine KUCs~(\tikzref{cifar100-mfmix-supergroup-limit}) from the first unbiased experiment, \cf \cref{fig:biased-vs-unbiased-bar}, serve as reference. Best strategies are indicated in the respective subtitle.
	}
	\label{fig:cifar100-mfmix-supergroup}%
\end{figure}

The \gls{aucroc} results for \naively generated off-the-shelf mixups are shown in \cref{fig:cifar100-mfmix-supergroup}.
Other metrics like the \gls{oscr} are anticipated here and can be found in the supplement for a more detailed analysis, this includes also the results for the C-EVM and \gls{pisvm}.

Similar to the findings in the first unbiased experiment in \cref{fig:biased-vs-unbiased-bar}, the \gls{dnn} (SPL) in this experiment also does not benefit from the additional background data, \cf \cref{subfig:cifar100-mfmix-supergroup-dnn}.
The \gls{aucroc} and \ccrfpr{10} of KvR show a steady increase.
Interestingly, this does not apply to the entire \gls{fpr} range, as the \ccrfpr{1} temporarily deteriorates.
This suggests that mixups may offer advantages at medium \glspl{fpr} while providing no benefit at low \glspl{fpr}, or vice versa.
This makes it difficult to make a universal statement, such as mixup always improves OSR performance.

The \gls{evm} in \cref{subfig:cifar100-mfmix-supergroup-evm} shows remarkable benefits from all strategies and mixup.
The AUC-ROC corresponds perfectly with \cref{tab:evm-toy-3class}.
SPL constricts the space for the KCs more than MPL, resulting in higher recognition rates of unknowns at the expense of knowns.
MPL and KvR loosen this constriction.
Not only the baseline, but also EVM (SPL) trained with genuine KUCs is outperformed.
This is an intriguing observation, showing that mixup improves \glspl{uuc} recognition to the same extent or even more than genuine dataset-related data, which is often expensive to obtain in real-world scenarios~\cite{wang2020focalmix}.
Although SPL is slightly better in the first experiment, \cf \cref{fig:biased-vs-unbiased-bar}, we recommend KvR as preferred strategy due to its consistently stable results across all \glspl{fpr}.

The result of the W-SVM is in \cref{subfig:cifar100-mfmix-supergroup-wsvm}.
Training with \gls{kvr} yields a comparable \gls{aucroc} gain to genuine \glspl{kuc}.
With a \mixknownratio of \num{0.1}, it exceeds the baseline by about \SIP{1.2}.

Improving \gls{osr} with mixup is not universally effective.
The \gls{osnn} and \gls{pisvm}, \cf supplement, decline in performance when mixup is applied.
A nearest neighbor based approach is particularly prone to unfavorable sample placement, which is exacerbated by \naive mixups.
We address this occupation problem in the next section.


\begin{figure*}[tb]
	\centering
	\includegraphicstikz{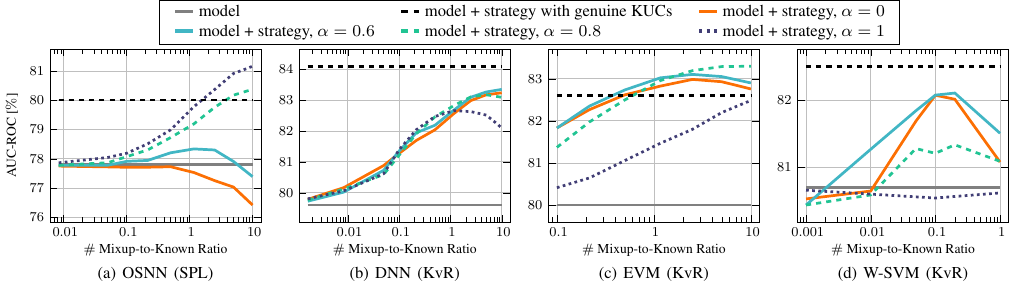}{.99\linewidth}%
	\myBelowcaptionskip%
	\caption{Unbiased results of the models exploiting constrained mixups on \cifar{100}.
	Shown is one strategy for each model with the \acrshort{aucroc} over the number of mixup samples.
	The baseline model~(\tikzref{cifar100-mfmix-constrained-supergroup-bl}) and the model exploiting genuine KUCs~(\tikzref{cifar100-mfmix-constrained-supergroup-limit}) serve as reference.}
	\label{fig:cifar100-mfmix-constrained-supergroup}%
\end{figure*}%

\subsection{Solving the occupation problem}\label{subsec:occupation}
Zhou \etal~\cite{zhou2021proser} mention the issue of already-occupied space between two \glspl{kc} by a third.
Since mixups are labeled as unknown, this can degrade the adjacent \glspl{kc}.
As augmentation technique or in feature learning, the occupation problem is not approached due to two reasons:
\begin{enumerate*}
	[label={\arabic*)}]
	\item Mixups are generated in mini-batches and checking on their location in the feature space in relation to \emph{all} other samples is infeasible.
	\item The occupation might be favorable as mixups push away the in-between class, increasing the space between all affected classes.
\end{enumerate*}
However, in case of pre-trained features, this can indeed impact \gls{osr} performance.
We hypothesize that it is the main reason for the \gls{osnn} failure case.

We propose an effective constraint for an on-the-fly mixup generation.
Let $\bar{\Bx}_y$ be a \gls{kc}'s centroid, calculated as mean of all samples within a class.
Using Euclidean distance $d(\cdot,\cdot)$, we determine the mean distance $\bar{d}$ among all class centroids.
A mixup sample is kept if its distance to all class centroids exceeds the scaled $\bar{d}$.
Scale $\alpha  =  0$ represents unconstrained generation and $\alpha  >  0$ promotes filtering:
\begin{equation}
	\label{eq:mixup-constraint}
	d(\tilde{\Bx},\bar{\Bx}_y) > \alpha \cdot \bar{d} \quad \forall y \in \MC_K \enspace \td
\end{equation}

We conduct the \cifar{100} experiment again, varying \mbox{$\alpha  \in  \{0, 0.6, 0.8, 1\}$}.
Results are presented in \cref{fig:cifar100-mfmix-constrained-supergroup} and we recommend to view the supplement for additional details.

The OSNN in \cref{subfig:cifar100-mfmix-constrained-supergroup-osnn} benefits from stronger constraints ($\alpha  \geq  0.8$), resulting in consistent \gls{aucroc} improvements of up to \SIP{3.4} compared to the baseline as the number of mixups increases.
The \ccrfpr{1} is erratic, but with FPR=\SIP{10}, it improves by almost \SIP{2}.
We note three observations:
\begin{enumerate*}
	[label={\arabic*)}]
	\item OSNN is prone to \naive mixups.
	\item Highly constrained mixups improve OSR, but not across the entire \gls{fpr} range.
	\item Performance deterioration is unlikely and can be mitigated by monitoring a held-out set.
\end{enumerate*}

The DNN (KvR) in \cref{subfig:cifar100-mfmix-constrained-supergroup-dnn} has little benefit with stronger restrictions, for $\alpha  =  1$ it even deteriorates after a \mixknownratio of $1$.
Mixups lead to an improvement of up to \SIP{4}, especially for FPRs $>$\SIP{1}.
We conclude:
\begin{enumerate*}
	[label={\arabic*)}]
	\item \Naive mixups enhance OSR for medium and larger FPRs.
	\item DNNs usually benefit from diverse data, constraining mixups might be counterproductive.
\end{enumerate*}

The EVM (KvR) in \cref{subfig:cifar100-mfmix-constrained-supergroup-evm} shows that no or lower constraints ($\alpha  <  0.8$) increase more with the number of mixup samples in the CCR than the more restricted ones.
The magnitude of the constraints affects different areas of the FPR.
Our conclusions are:
\begin{enumerate*}
	[label={\arabic*)}]
	\item Due to the variable behavior of different constraints, the EVM can be optimized for a desired FPR.
	\item Finding the optimal configuration is complex and may require a hyperparameter search.
\end{enumerate*}

While the \gls{aucroc} of the \gls{wsvm} in \cref{subfig:cifar100-mfmix-constrained-supergroup-wsvm} appears promising, the \gls{oscr} is irregular with a downward trend.
This behaviour is underlined by the toy example in the supplement, where different \mixknownratios lead to erratic changes at low confidences.

%% file: tables/evm-toy-3class-vertical.tex
\begin{table}[tb]
	\centering
	\footnotesize
	\setlength{\tabcolsep}{1.2pt}
	\renewcommand{\arraystretch}{0.5}
	\renewcommand{\mytmp}{0.118}%
	\setlength{\belowcaptionskip}{2pt}
	\caption{EVM class boundaries with the baseline in the top left and applied strategies.
	This toy dataset contains dark-edged dots from $3$ KCs and orange dots as mixups.
	Colored areas display class assignment, with opacity indicating confidence, where white is zero confidence or, conversely, high confidence for open space.}\label{tab:evm-toy-3class}%
	\begin{tabular}{cc >{\centering\arraybackslash} m{\mytmp\textwidth} >{\centering\arraybackslash} m{\mytmp\textwidth} >{\centering\arraybackslash} m{\mytmp\textwidth}}
		\toprule\addlinespace
		\multicolumn{2}{c}{\multirow{2}{*}[1.25ex]{\includegraphics[width=0.1\textwidth]{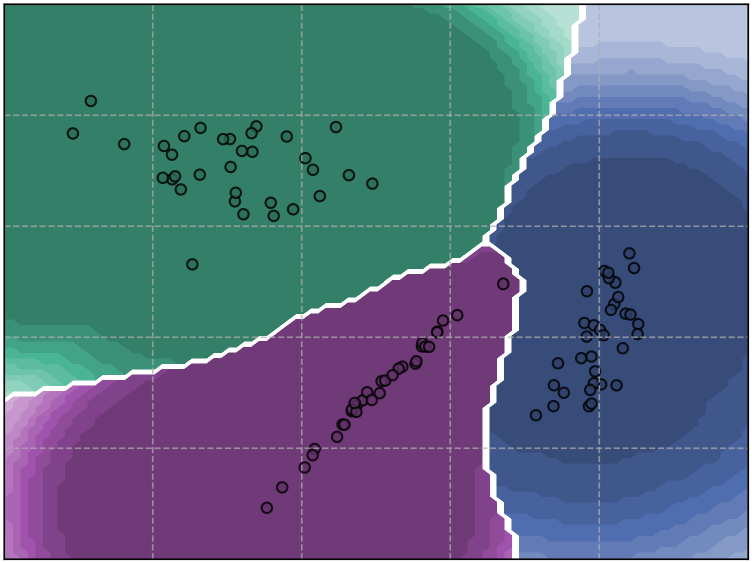}}} & \multicolumn{3}{c}{\textbf{Strategy}} \\\addlinespace
		\cmidrule{3-5}\addlinespace
		& & \textbf{EVM (SPL)} & \textbf{EVM (MPL)} & \textbf{EVM (KvR)} \\\addlinespace
		\multirow{2}{*}{\rotatebox[origin=c]{90}{\makecell{\textbf{Mixup-to-} \\\textbf{Known Ratio}}}} & \num{0.1} &
		\includegraphics[width=\mytmp\textwidth]{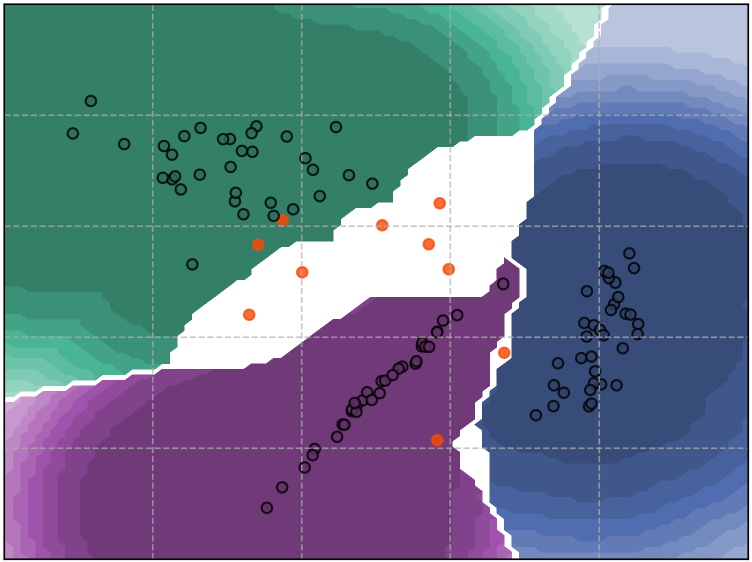} &
		\includegraphics[width=\mytmp\textwidth]{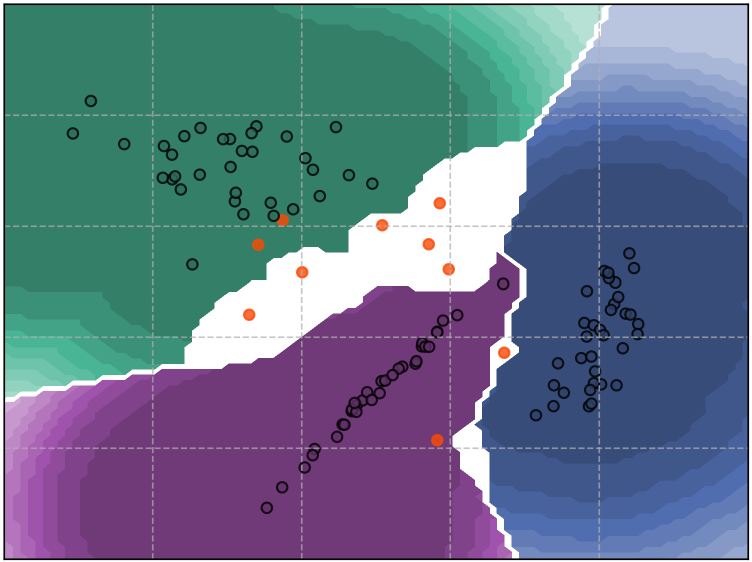} &
		\includegraphics[width=\mytmp\textwidth]{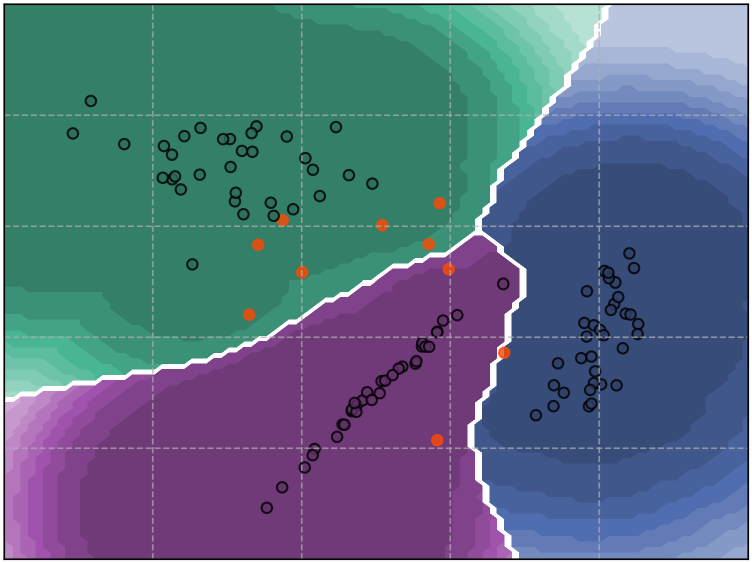} \\
		& \num{10} &
		\includegraphics[width=\mytmp\textwidth]{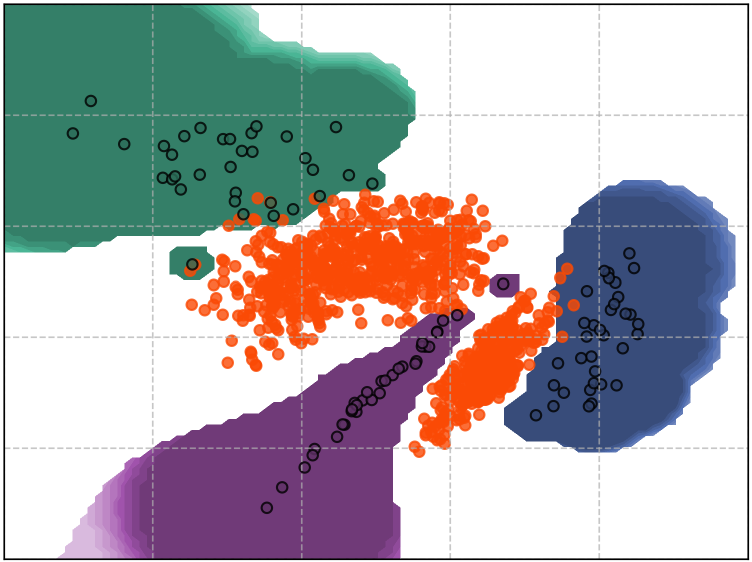} &
		\includegraphics[width=\mytmp\textwidth]{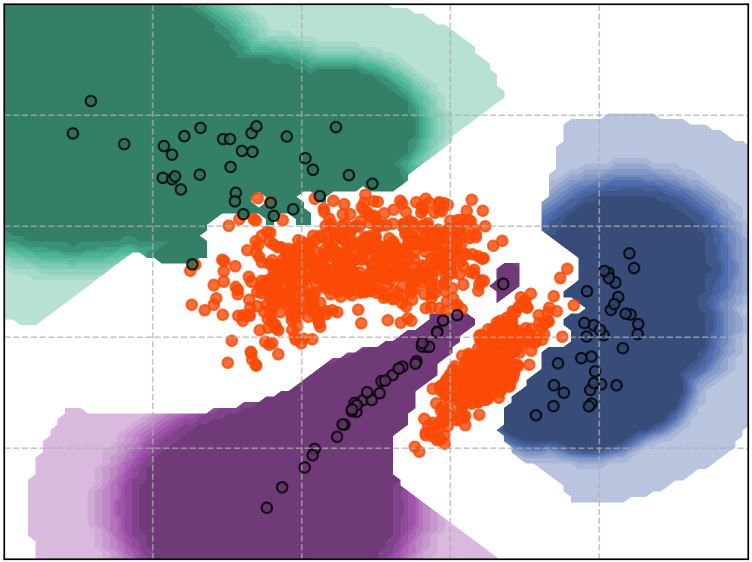} &
		\includegraphics[width=\mytmp\textwidth]{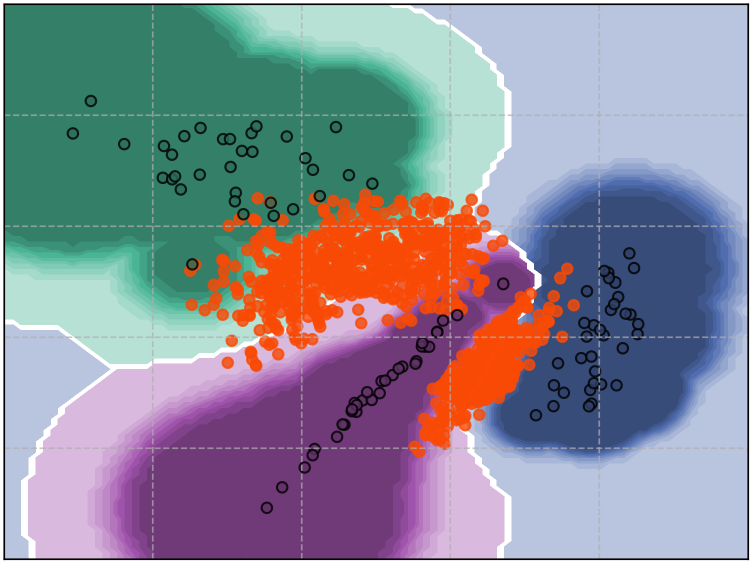} \\
		\bottomrule
	\end{tabular}
	\vspace{-0.75\baselineskip}
\end{table}%

%% file: 06discussion.tex
\section{Discussion and Conclusion}\label{sec:discussion}

This paper presents LORD to explicitly model open space by exploiting \glspl{kuc} for open-set learning and to boost \glspl{osr} models.
Our key findings are as follows:

\paragraph{Known unknowns improve OSR learning.}

Exploiting \glspl{kuc} to train \gls{osr} models improves over baselines ignoring such samples.
Considering \glspl{kuc} improves the detection of \glspl{kuc} and \glspl{uuc}.
The performance for \glspl{uuc} is usually lower.
Practitioners should carefully select the evaluation strategy and only benchmark with \glspl{kuc} if this reflects the conditions in the addressed use case.

\paragraph{The training strategy matters.}
\Gls{spl} can be considered as a baseline due to its simplicity.
When used with synthetic \glspl{kuc}, \gls{spl} is outperformed by competing methods.
\Gls{mpl} shows comparable results but is not tractable for every model, treating each \gls{kuc} sample as a separate class.
\Gls{kvr} outperforms the other strategies in most benchmarks and requires fewer parameters to learn.

\paragraph{The OSR evaluation measure matters.}
\Gls{aucroc} measures the ability to distinguish knowns and unknowns, but relying solely on this metric can create false expectations.
The relevant FPR range for OSR is usually $<$\SIP{10}, occupying a small part of the AUC and being outweighed by the range $>$\SIP{10}.
\Gls{aucroc} improvements might be due to enhancements at high \glspl{fpr}, which have limited relevance for \gls{osr}.
It is crucial to report \gls{tpr} or \gls{ccr} at low \glspl{fpr}.
For future work, the OpenAUC~\cite{wang2022openauc} could also be a helpful alternative to measure \gls{osr} performance.

\paragraph{Mixups can surrogate genuine known unknowns.}
Mixups serve as a lightweight and effective surrogate for \glspl{kuc}.
\Gls{dnn}, \gls{evm}, and \gls{osnn} benefit the most from mixups.
\Gls{osnn} requires extra filtering to mitigate the occupation problem.
Filtering does not consistently improve all models but increases the \gls{ccr} in certain \gls{fpr} ranges.
The aim of solving the occupation problem is to maintain a high \gls{kc} classification.
A suitable compromise between unknown detection and known recognition has to be found.

\paragraph{LORD is extendable to open-world learning.}

Future work refines the generation of known unknowns.
The constraints to address the occupation problem shall be enhanced.
The highly efficient mixup synthesis proposed here is suitable for open-world learning and future experiments should involve open-world capable classifiers.

%% file: 10supmat.tex
\appendix
\renewcommand\thefigure{\thesection.\arabic{figure}}
\renewcommand\thetable{\thesection.\arabic{table}}
\setcounter{figure}{0}
\setcounter{table}{0}

\section{Appendix}\label{ch:appendix}
This supplementary material is organized as follows:
\begin{itemize}
	\item \Cref{sec:deployed-strategies-for-open-set-models-details} provides some further details on the implementation of the strategies in the models.
	\item \Cref{sec:performance-of-meta-learning-strategies} presents the results from the \gls{cevm} and \gls{pisvm}~\cite{jain2014pisvm}, exploiting genuine \glspl{kuc} with the learning strategies.
	\item \Cref{sec:augmenting-open-set-models-by-manifold-mixup} shows visualizations of the decision boundaries on a toy dataset for all models and various \mixknownratios.
	It also includes additional \gls{osr} measures for the models in the main work, the C-EVM, and the \gls{pisvm}, using mixups as \gls{kuc} surrogates.\phantom{\gls{evm}}
	\item \Cref{sec:solving-the-occupation-problem} contains additional metrics for the models in the main work, the C-EVM, and the \gls{pisvm}, with constrained mixups.\phantom{\gls{svm}}
\end{itemize}

\subsection{Deployed strategies for open-set models --\\additional details}\label{sec:deployed-strategies-for-open-set-models-details}
In this section, we outline how to deploy the strategies to \num{6} different \gls{osr} models from the following \num{4} categories.

\glsreset{osnn}

\paragraph{\Gls{osnn}.}
The \gls{osnn}~\cite{junior2017osnn} exploits the ratio between the two nearest samples from distinct classes as confidence value.
Let $d_i$ and $d_j$ be Euclidean distances between a query sample $\Bx$ and its two closest training samples, $\Bx_i$ and $\Bx_j$, where $y_i \!\neq\! y_j$.
The ratio is computed as $r = \sfrac{d_i}{d_j}$ with $d_i < d_j$.
If $r \leq \delta$, query $\Bx$ is labeled as $y_i$, otherwise it is labeled as $u$.

\gls{spl} includes all \glspl{kuc} in distance ratio computation and class prediction.
\Gls{mpl} treats each \gls{kuc} as a separate class, impacting the reject option only.
For \gls{kvr}, the distance ratio is set to the maximum $r \!=\! 1$ if the nearest sample $\Bx_i$ belongs to a \gls{kuc}.
Thus, only \glspl{kc} are used for label prediction, while both \glspl{kc} and \glspl{kuc} determine distance ratios.
The strategies primarily vary in the confidence computation, leading to nearly identical results.

\glsreset{dnn}

\paragraph{\Gls{dnn}.}
Given the vast possibilities of using \glspl{kuc} in training \glspl{dnn} by tailoring losses, we opt for the classical cross-entropy loss.
The feature extractors mentioned in the dataset-specific paragraphs of the main manuscript are used.
A single fully-connected layer with softmax activation is attached and finetuned.

For \gls{spl}, we expand the number of output units by one class.
We do not evaluate \gls{mpl} as it does not scale to large \glspl{kuc} sets.
For \gls{kvr}, we note that this strategy is equivalent to the entropic open-set loss~\cite{dhamija2018oscr}.
The objective of entropic open-set is to predict a uniform distribution of unknowns, while \glspl{kc} are learned according to a cross-entropy loss.

\glsreset{evm}

\paragraph{\Glspl{evm}.}
The \gls{evm}~\cite{rudd2017evm}
estimates a Weibull distribution for each sample, considering distances to the nearest samples from other classes.
It implements an \gls{ovr} scheme as it uses distances to \emph{rest}-class samples for each \emph{one}-class sample.
The $\tau$ smallest distances termed \emph{tail} are used to estimate Weibull distributions.

\Gls{spl} and \gls{mpl} use $1$ or $|\MT_u|$ pseudo-classes, respectively.
Unlike \gls{spl}, \gls{mpl} allows \glspl{kuc} in the tail of other \glspl{kuc}, causing regularization among close \glspl{kuc}.
This effect is visible when comparing SPL and MPL in \cref{tab:evm-cevm-toy-3class}, where the space between \glspl{kc} and densely populated \gls{kuc} areas is prioritized for \glspl{kc}.
\Gls{kvr} uses \glspl{kuc} only in the tails of \glspl{kc}, and \glspl{kuc} never act as one-class themselves, leading to gentle transitions between decision boundaries.

We also explore the C-EVM~\cite{henrydoss2020cevm}, which employs DBSCAN clustering on a per-class basis before the EVM fitting.
Clustering calculates centroids for each cluster, serving as proxies for all samples within the cluster.
EVM fitting is exclusively performed on these centroids.
This model-agnostic preprocessing technique can be applied to any other method mentioned in this work.
One particular aspect of interest is whether clustering can counteract the label noise of the MPL strategy.
For SPL and MPL, clustering is applied to the entire \glspl{kuc} as a unified class.
For MPL, the resulting class centroids are treated as independent classes again, replacing redundant \glspl{kuc} and avoiding label noise.
For KvR learning, only the \glspl{kc} undergo cluster-based reduction while leaving the \glspl{kuc} unaffected.

\glsreset{svm}
\glsreset{pisvm}
\glsreset{wsvm}

\paragraph{\Glspl{svm}.}
We deploy the training strategies to two \gls{svm} variants.
The \gls{wsvm}~\cite{scheirer2014wsvm} combines a one-class \gls{svm} and a binary \gls{ovr} \gls{svm} for each class.
Weibull distributions are estimated from both \glspl{svm} and probabilities are determined by these Weibull distributions.
The \gls{pisvm} ~\cite{jain2014pisvm} predicts unnormalized posterior inclusion probabilities with \acrshort{rbf} kernels.
Weibull distributions are estimated on samples near decision boundaries.

The training strategies \gls{spl} and \gls{mpl} differ only in the number of pseudo-classes.
We omit \gls{mpl} due to its limited scalability to large number of classes.
Also, \gls{mpl} is makes it challenging to serve the \glspl{svm}' requirement of at least three samples per class (preferably more).
For the same reason, we do not evaluate the \glspl{svm} on \gls{lfw}.
The KvR strategy can be deployed to the SVMs by not representing the KUCs as a positive one-class.
Instead, they are always considered part of the rest-class during training.

\vspace{2\baselineskip}

\newpage

\subsection{Performance of training strategies --\\additional results}\label{sec:performance-of-meta-learning-strategies}
This section complements the experiments in which genuine \glspl{kuc} are exploited by the three learning strategies:
\begin{enumerate*}
	[label={\arabic*)}]
	\item \gls{spl},
	\item \gls{mpl}, and
	\item \gls{kvr}.
\end{enumerate*}

\Crefrange{subfig:genuine-kuc-supmat-cevm-cifar}{subfig:genuine-kuc-supmat-cevm-casia} show the biased \gls{oscr} of the C-EVM within the open-set relevant \gls{fpr} range on \cifar{100}, \gls{lfw}, and Tiny \gls{cwf}.
The C-EVM demonstrates similar behavior to the vanilla \gls{evm}~\cite{rudd2017evm}.
All strategies perform comparably well and outperform the baseline.
However, it remains inconclusive whether the prior clustering of background data in SPL and MPL prevents potential label noise, leading to improved detection.
Two possible conclusions arise:
\begin{enumerate*}
	[label={\arabic*)}]
	\item In all three datasets, there is no noise within the genuine \glspl{kuc}.
	\item The C-EVM performs well even without prior clustering of background data, as in \gls{kvr}.
\end{enumerate*}

\Cref{subfig:genuine-kuc-supmat-pisvm-cifar,subfig:genuine-kuc-supmat-pisvm-casia} show the biased results of the \gls{pisvm}.
For \cifar{100}, consistent improvement over the baseline is evident.
However, for Tiny \gls{cwf}, SPL at FPRs greater than \SIP{1} results in a degradation of the \gls{ccr}.
The \gls{pisvm} appears to encounter challenges in modeling all \glspl{kuc} within a single class, whereas the \gls{wsvm} in the main work performs well.
This discrepancy might be attributed to the \gls{wsvm}'s use of a one-class and a binary SVM for each class, while the \gls{pisvm} deals solely with a binary SVM.

\begin{figure}[tb]
	\centering
	\includegraphicstikz{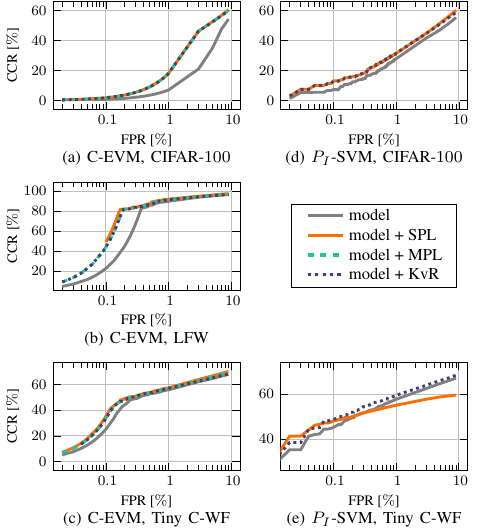}{\linewidth}%
	\myBelowcaptionskip%
	\caption{Results of the biased evaluation of $2$~models (column-wise) exploiting genuine KUCs with the strategies and the baseline on $3$ datasets (row-wise).
	The models are C-EVM in~\csubrefrange{subfig:genuine-kuc-supmat-cevm-cifar}{subfig:genuine-kuc-supmat-cevm-casia} and \acrshort{pisvm} in~\subref{subfig:genuine-kuc-supmat-pisvm-cifar} and \subref{subfig:genuine-kuc-supmat-pisvm-casia}.
	Shown is the biased \acrfull{oscr} in the open-set relevant FPR range up to \SIP{10}.
	}
	\label{fig:genuine-kuc-supmat}%
	\vspace{25\baselineskip}
\end{figure}%

\input{tables/osnn-dnn-toy-3class.tex}

\newpage

\subsection{Augmenting models by manifold mixup --\\toy examples and additional results}\label{sec:augmenting-open-set-models-by-manifold-mixup}
In this section, we present additional decision boundary illustrations using a toy example and various \mixknownratios.
The second paragraph contains additional open-set measures for the experiment involving \naive mixup samples as \glspl{kuc} surrogates.

\paragraph{Toy example visualizations.}
\Cref{tab:osnn-dnn-toy-3class} shows the behavior of the \gls{osnn} and \gls{dnn} with the deployed strategies.
As observed in the main manuscript, the strategies exhibit minimal differences in OSNN.
The model only considers two nearest neighbors, leaving limited scope for variation.

For the DNN, the open space expands as the number of mixups increases, consequently pushing the decision boundaries closer to the known classes.
In instances where SPL encounters unfavorably located classes, such as the purple one, the outcome can be very unfavorable.
In contrast, the KvR approach is more lenient and consistently offers more flexibility by employing an appropriate threshold.

\Cref{tab:evm-cevm-toy-3class} shows both EVM variants with the toy example.
As the vanilla EVM has already been discussed in the main manuscript, this section displays the remaining \mixknownratios.

Unlike other methods, the training data for the C-EVM is shown here after the cluster-based reduction.
As described in \cref{sec:deployed-strategies-for-open-set-models-details}, SPL and MPL treat \glspl{kuc} as a single class during clustering, potentially reducing label noise in MPL.
In KvR, the background data is not reduced.
Notably, we observe a concave-like function when combining oversampling mixups with cluster-based reduction.
Oversampling generates larger coherent clusters, leading to a decrease in the number of clusters beyond a certain point.
Each cluster is reduced to a centroid, resulting in the observation that more mixups lead to fewer mixups in this type of reduction.

\Cref{tab:svms-toy-3class} displays the toy example alongside the \gls{wsvm} and \gls{pisvm}.
The decision boundaries of both \glspl{svm} are generally comparable with only slight variations.
In the low-confidence range, there are occasional abrupt changes.
However, since the nearly transparent areas correspond to very low confidence values, a suitable threshold would usually consider these areas as open space.

\vspace{4\baselineskip}

\input{tables/evm-cevm-toy-3class.tex}

\phantom{foo}

\vspace{10\baselineskip}

\phantom{foo}

\vspace{10\baselineskip}

\input{tables/svms-toy-3class.tex}

\begin{figure*}[tb]
	\centering
	\includegraphicstikz{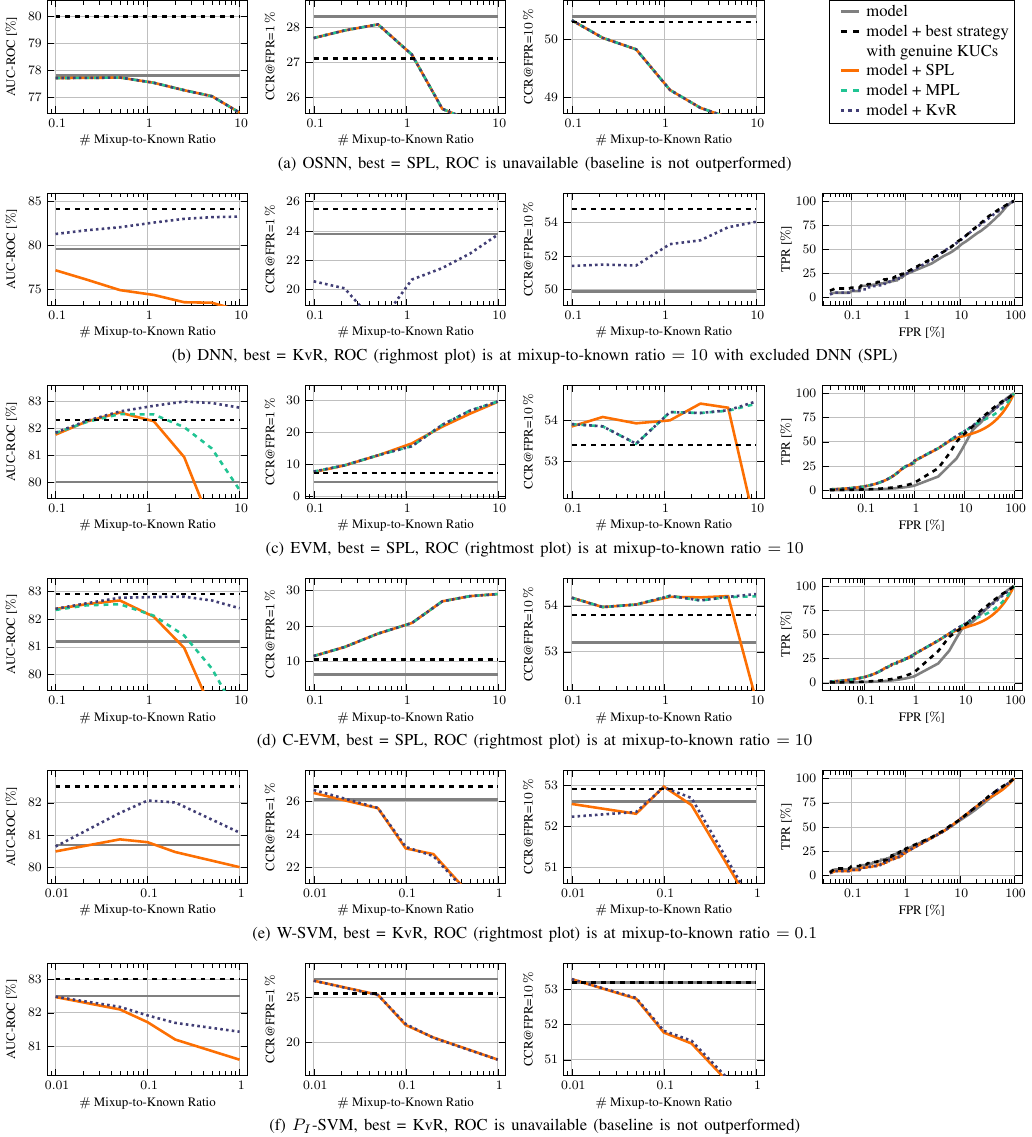}{\linewidth}%
	\caption{Unbiased results of all models (row-wise) trained with the different strategies and mixup samples on \cifar{100}.
	The metrics are (left to right): \acrshort{aucroc}, \ccrfpr{1}, \ccrfpr{10}, and the \acrshort{roc}. The first $3$ metrics are shown \wrt the \mixknownratio. The ROC is depicted at a specific \mixknownratio. The baseline model without KUCs~(\tikzref{cifar100-mfmix-supmat-bl}) and the best strategy of each model trained with genuine KUCs~(\tikzref{cifar100-mfmix-supmat-limit}) from the first unbiased experiment, \cf \cref{fig:biased-vs-unbiased-bar} in the main work, serve as reference. This best strategy and the \mixknownratio of the ROC are indicated in the respective subtitles.}
	\label{fig:cifar100-mfmix-supmat}%
\end{figure*}

\paragraph{Additional results.}
This paragraph completes the experiments involving the replacement of genuine \glspl{kuc} with mixup samples.
\Cref{fig:cifar100-mfmix-supmat} combines additional open-set measures for the methods discussed in the main manuscript, along with the remaining results for the C-EVM and \gls{pisvm}.
The extended evaluation includes the \gls{aucroc}, \ccrfpr{1}, \ccrfpr{10}, and the \gls{roc} at a specific \mixknownratio.
The latter represents the point of optimal performance while exploiting mixups.
The optimal point is indicated in the respective subtitle of each model.

In \cref{subfig:cifar100-mfmix-supmat-osnn}, \gls{osnn} demonstrates a failure case, exhibiting a degradation in performance across all metrics.
However, we demonstrate that resolving the occupation problem enhances \glspl{osr} performance for this classifier.

In \cref{subfig:cifar100-mfmix-supmat-dnn}, KvR demonstrates an improvement in the \gls{aucroc}, but this improvement comes partly at the expense of the \ccrfpr{1}, which decreases to the \mixknownratio of \num{0.3} and then starts to increases again.
In comparison, \ccrfpr{10} steadily increases and reaches \SIP{54} at a \mixknownratio of $10$, while the baseline achieves \SIP{50}.
This gain is also evident in the \gls{roc} where it extends to very high FPRs.
In conclusion, mixup proves to be a suitable approach for improving \gls{dnn} (\gls{kvr}) in applications with medium security requirements.

In \cref{subfig:cifar100-mfmix-supmat-evm,subfig:cifar100-mfmix-supmat-cevm}, the \gls{evm} and C-EVM outperform the baseline by exploiting mixups.
Both variants show similar behavior, with \gls{kvr} outperforming the other strategies in this experiment.
The most significant improvement over the baseline is observed at the \ccrfpr{1}, reaching over \SIP{25} at the highest evaluated \mixknownratio.
This advantage also extends to the \ccrfpr{10}, except for \gls{spl}, which experiences a sharp drop with many mixups, as previously observed in the \gls{aucroc}.
The \gls{roc} shows that for \gls{spl} and \gls{mpl}, the \gls{tpr} deteriorates at \glspl{fpr} greater than \SIP{10}.
This suggests that mixups have no positive effect on the closed-set case.
However, there is a tremendous gain at \glspl{fpr} between \SIrange{0.1}{10}{\percent}.

While the \gls{wsvm} in \cref{subfig:cifar100-mfmix-supmat-wsvm} temporarily outperforms the baseline, the \gls{pisvm} in \cref{subfig:cifar100-mfmix-supmat-pisvm} does not benefit from the mixup samples.
It is noteworthy that training with genuine \glspl{kuc} also did not results in any enhancement.
A reasonable attempt at improvement would involve conducting a hyperparameter search for training with \glspl{kuc} as well.

\begin{figure*}[tb]
	\centering
	\includegraphicstikz{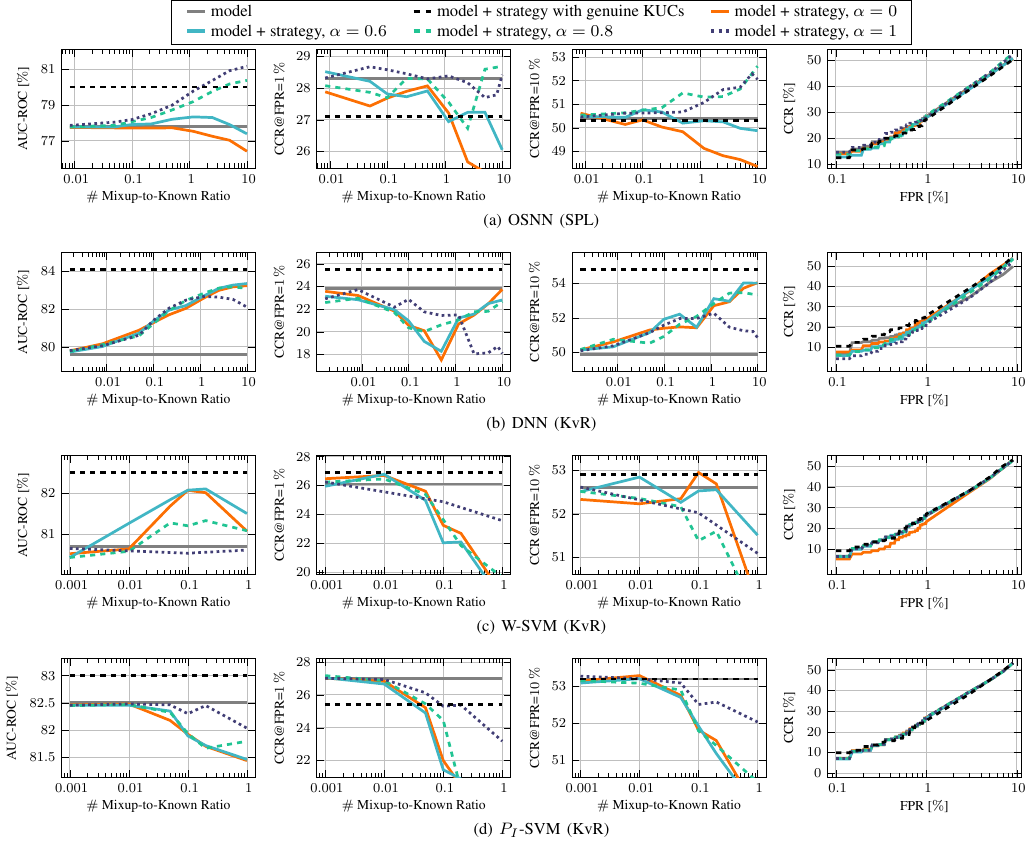}{\linewidth}%
	\caption{Unbiased results of $4$ models (row-wise) exploiting constrained mixups on \cifar{100}. The metrics are (left to right): AUC-ROC, \ccrfpr{1}, \ccrfpr{10}, and the OSCR at a certain \mixknownratio. The OSCR corresponds to the maximum of each curve in the \ccrfpr{10}. For example, in \subref{subfig:cifar100-mfmix-constrained-supmat-osnn}, for $\alpha=0.8$ it is the \mixknownratio of $10$ and for $\alpha=1$ the ratio of $8$.}
	\label{fig:cifar100-mfmix-constrained-supmat-osnn-dsnn-svms}%
\end{figure*}%

\subsection{Solving the occupation problem --\\additional results}\label{sec:solving-the-occupation-problem}
\begin{figure*}[tb]
	\centering
	\includegraphicstikz{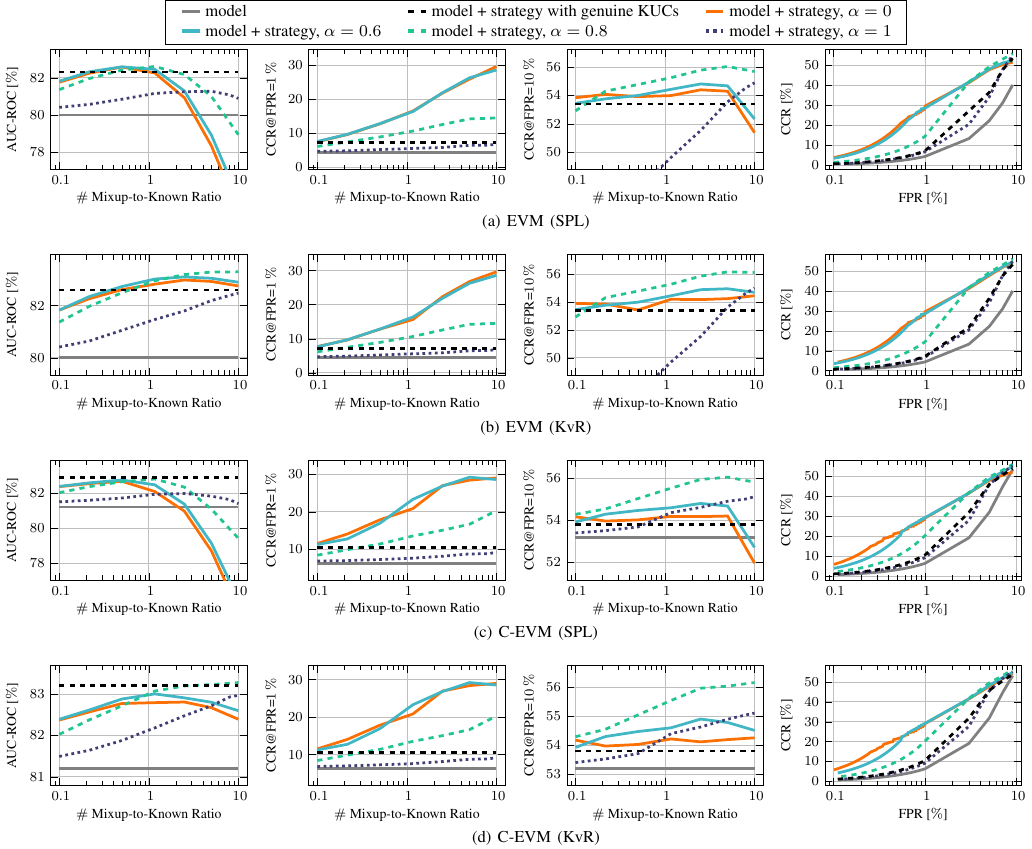}{\linewidth}%
	\caption{Unbiased results of the EVM variants with SPL and KvR (row-wise) exploiting constrained mixups on \cifar{100}. The metrics are (left to right): AUC-ROC, \ccrfpr{1}, \ccrfpr{10}, and the OSCR at a specific \mixknownratio. The OSCR corresponds always to the maximum of each curve at the \ccrfpr{1}.}
	\label{fig:cifar100-mfmix-constrained-supmat-evms}%
\end{figure*}

This section contains additional results of the assessment with constrained mixups to solve the occupation problem.
While the main manuscript focuses on the \gls{aucroc}, here we provide the other open-set measures as well.
\Cref{fig:cifar100-mfmix-constrained-supmat-osnn-dsnn-svms} displays the results for the \gls{osnn} (SPL), \gls{dnn} (KvR), \gls{wsvm} (KvR), and \gls{pisvm}.
\Cref{fig:cifar100-mfmix-constrained-supmat-evms} contains the results for the \gls{evm} and C-EVM, both with SPL and KvR.

In general, the \gls{osnn} in \cref{subfig:cifar100-mfmix-constrained-supmat-osnn} benefits from more constrained mixups.
While the \ccrfpr{1} varies unstably, the \ccrfpr{10}  shows improvement.
In contrast, the \gls{dnn} (KvR) in \cref{subfig:cifar100-mfmix-constrained-supmat-dnn} does not show any improvements with constrained mixups.
Stronger constraints can reduce the performance drop in the \ccrfpr{1} by \SIP{3}, but with $\alpha = 1$ it appears merely shifted.
Both \gls{svm} variants in \cref{subfig:cifar100-mfmix-constrained-supmat-wsvm,subfig:cifar100-mfmix-constrained-supmat-pisvm} marginally benefit from stronger constraints.
Their overall downward trend is reduced with $\alpha=1$ and could potentially be further improved with even stronger constraints.
However, based on the current results, the exploitation of mixup, or \glspl{kuc} in general, in combination with \glspl{svm} is limited for the detection of \glspl{uuc}.

The EVM variants in \cref{fig:cifar100-mfmix-constrained-supmat-evms} exhibit minimal variation.
Lower constraints enhance the \ccrfpr{1} while a constraint with $\alpha=0.8$ promotes the \ccrfpr{10}.
This difference in behavior can be leveraged when considering different safety requirements focused on different FPRs.

%% file: tables/osnn-dnn-toy-3class.tex
\begin{table*}[tb]
	\centering
	\setlength{\tabcolsep}{1.1pt}
	\renewcommand{\arraystretch}{0.5}
	\renewcommand{\mytmp}{0.17}%
	\caption{OSNN (top) and DNN (bottom) class boundaries with applied strategies and a column-wise increase in mixup samples.
	This toy dataset contains dark-edged dots from $3$ \acrfullpl{kc} and orange dots as mixups.
	Colored areas display class assignment, with opacity indicating confidence, where white is zero confidence or, conversely, high confidence for open space.}\label{tab:osnn-dnn-toy-3class}%
	\begin{tabular}{l >{\centering\arraybackslash} m{\mytmp\textwidth} >{\centering\arraybackslash} m{\mytmp\textwidth} >{\centering\arraybackslash} m{\mytmp\textwidth} >{\centering\arraybackslash} m{\mytmp\textwidth} >{\centering\arraybackslash} m{\mytmp\textwidth}}
		\toprule
		\textbf{Strategy} & \multicolumn{4}{c}{\textbf{$\bm{\#}$ \MixKnownRatio}} \\
		\cmidrule{2-6}
		& \num{0} & \num{0.1} & \num{0.5} & \num{1} & \num{10} \\
		\midrule
		\makecell{\textbf{OSNN (SPL)} \\\textbf{\&}\\\textbf{OSNN(MPL)}} &
		\includegraphics[width=\mytmp\textwidth]{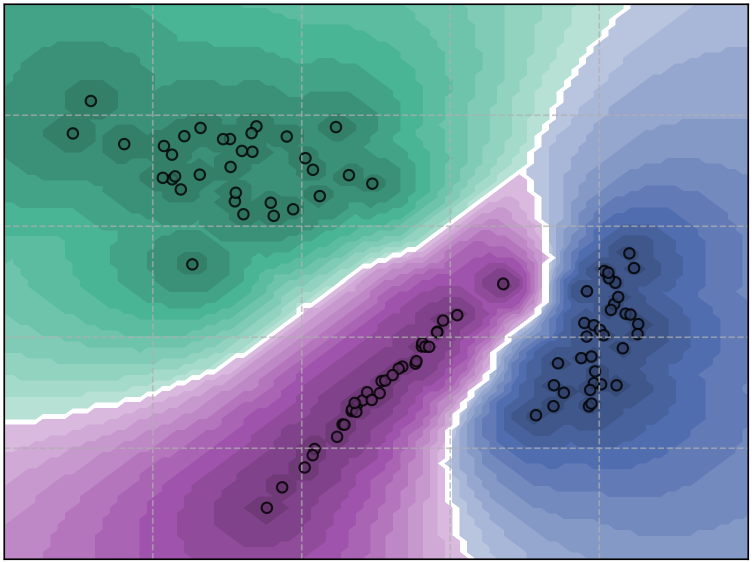} &
		\includegraphics[width=\mytmp\textwidth]{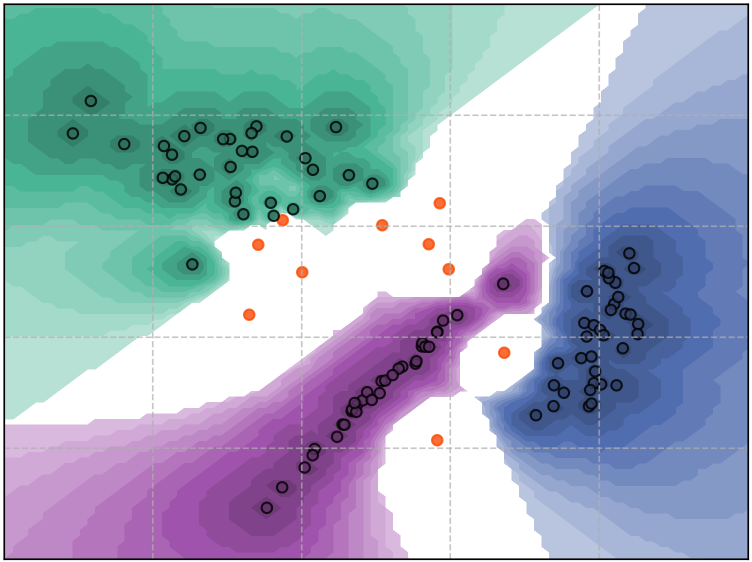} &
		\includegraphics[width=\mytmp\textwidth]{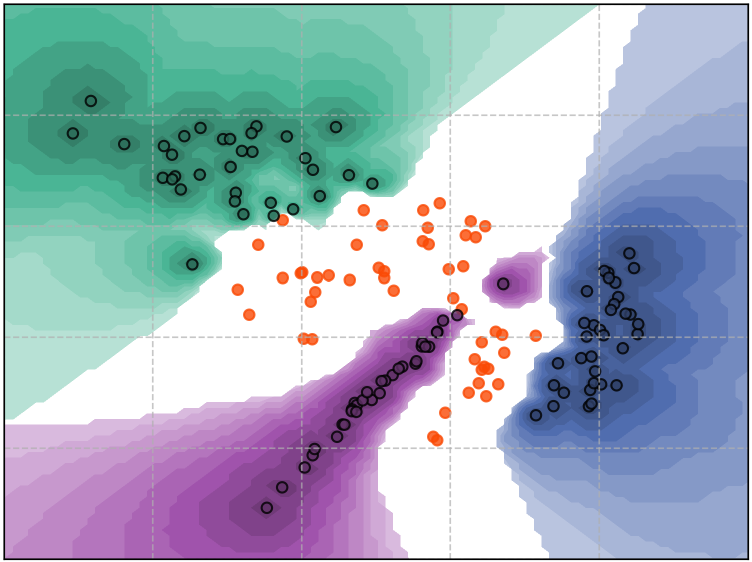} &
		\includegraphics[width=\mytmp\textwidth]{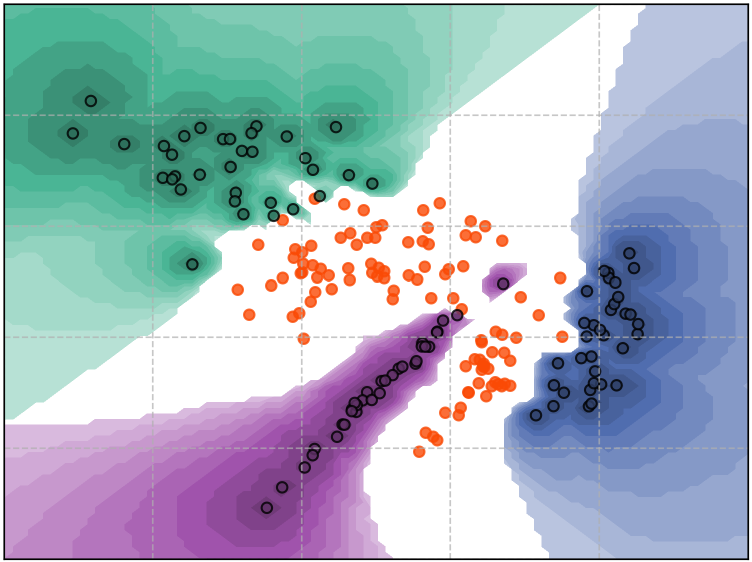} &
		\includegraphics[width=\mytmp\textwidth]{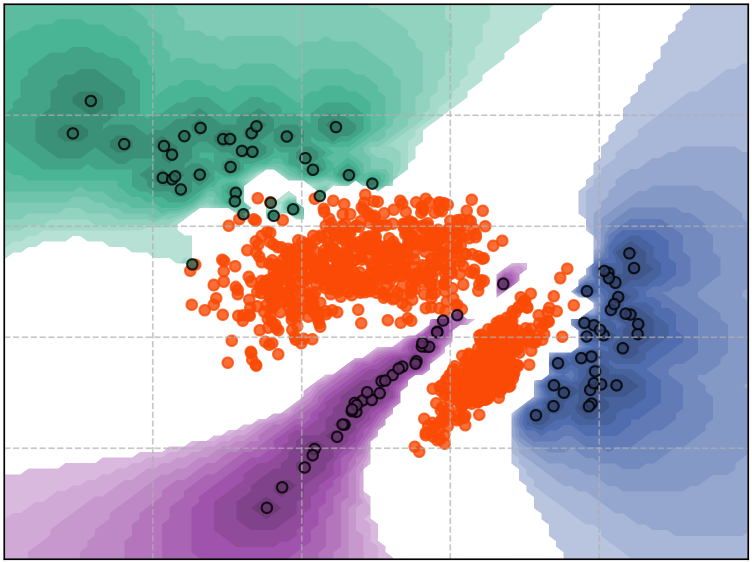} \\
		\textbf{OSNN (KvR)} &
		\includegraphics[width=\mytmp\textwidth]{figures/toy-example/ratio/osnn/OSNN-ignore-0} &
		\includegraphics[width=\mytmp\textwidth]{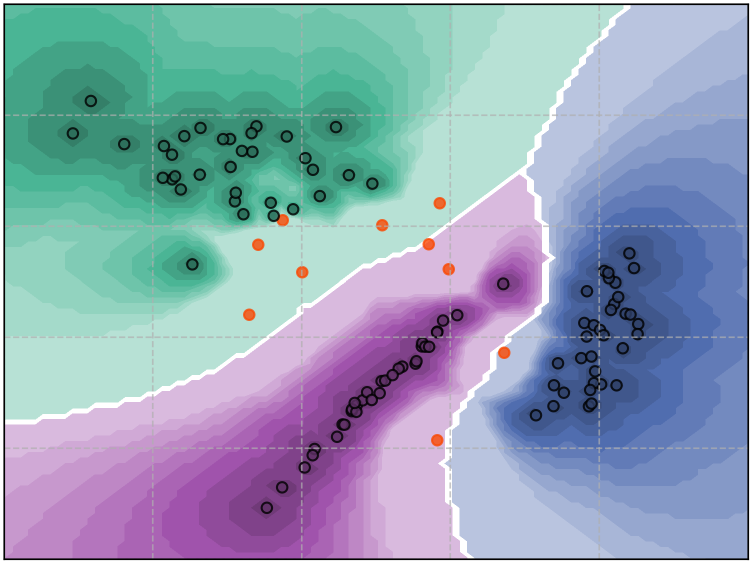} &
		\includegraphics[width=\mytmp\textwidth]{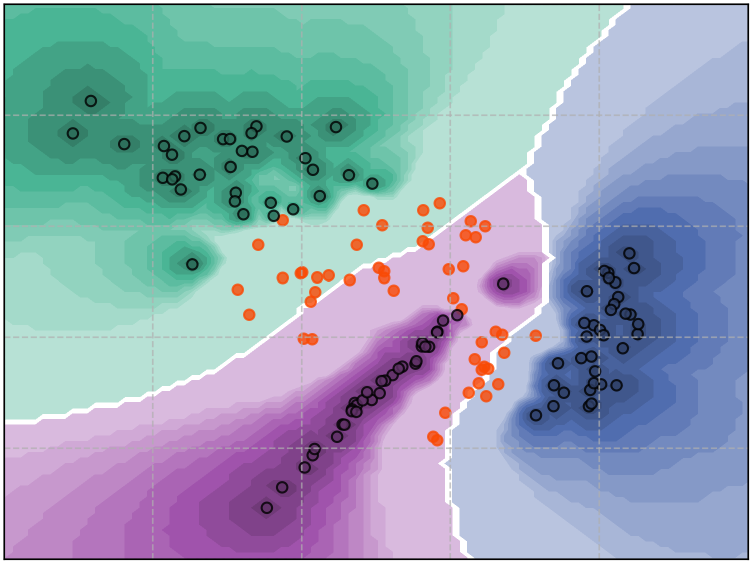} &
		\includegraphics[width=\mytmp\textwidth]{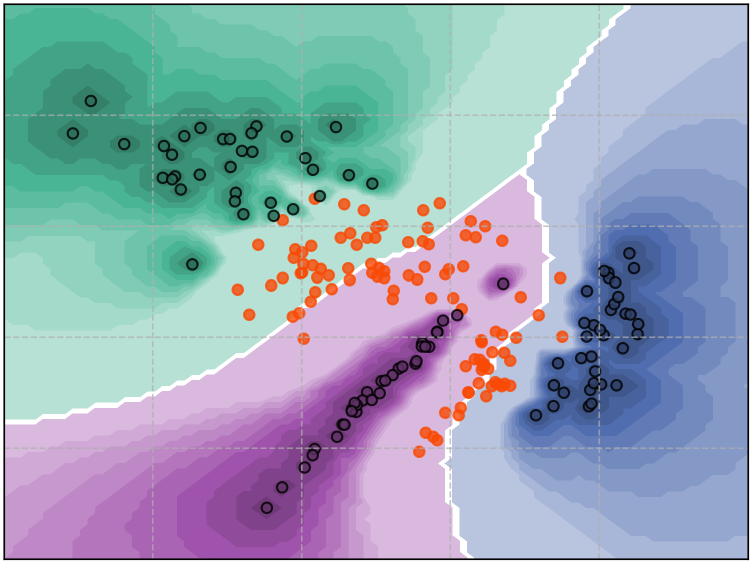} &
		\includegraphics[width=\mytmp\textwidth]{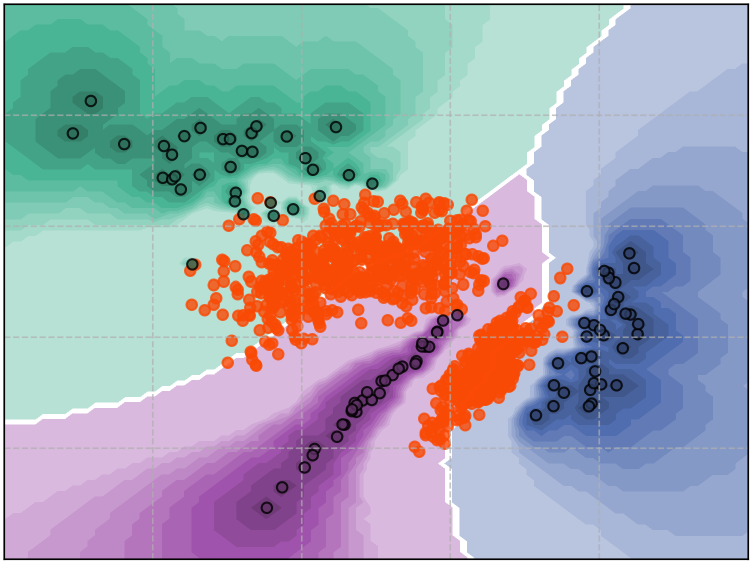} \\
		\midrule
		\textbf{DNN (SPL)} &
		\includegraphics[width=\mytmp\textwidth]{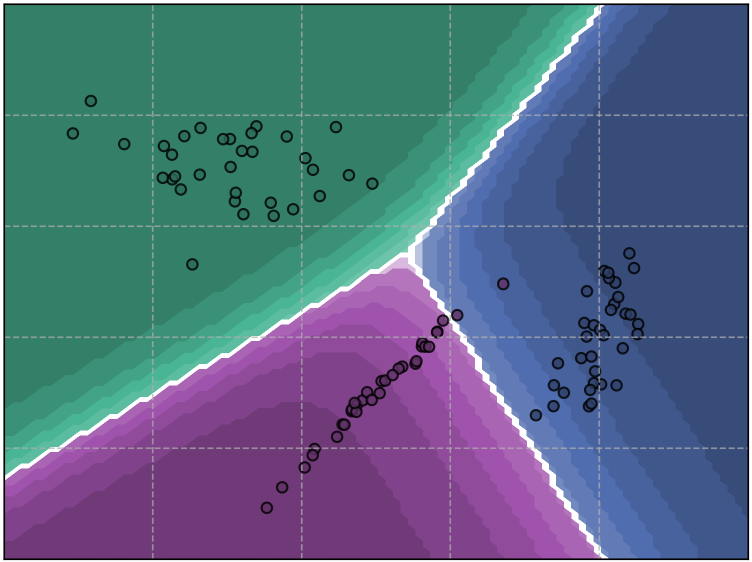} &
		\includegraphics[width=\mytmp\textwidth]{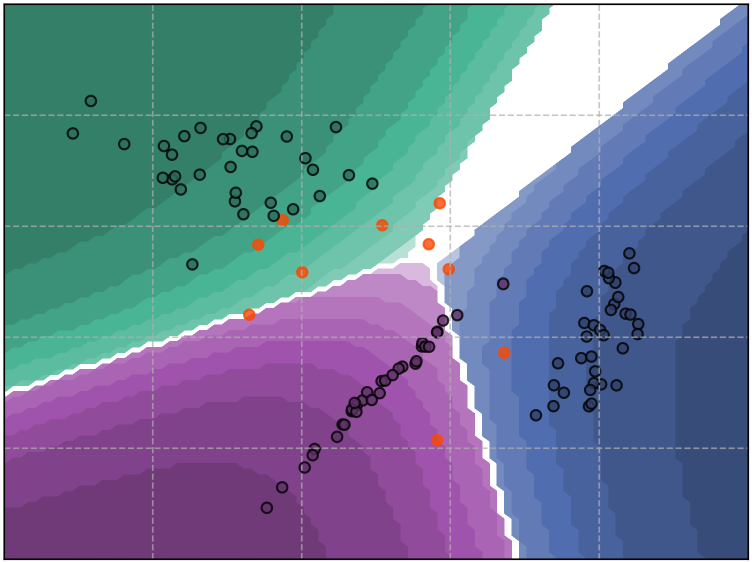} &
		\includegraphics[width=\mytmp\textwidth]{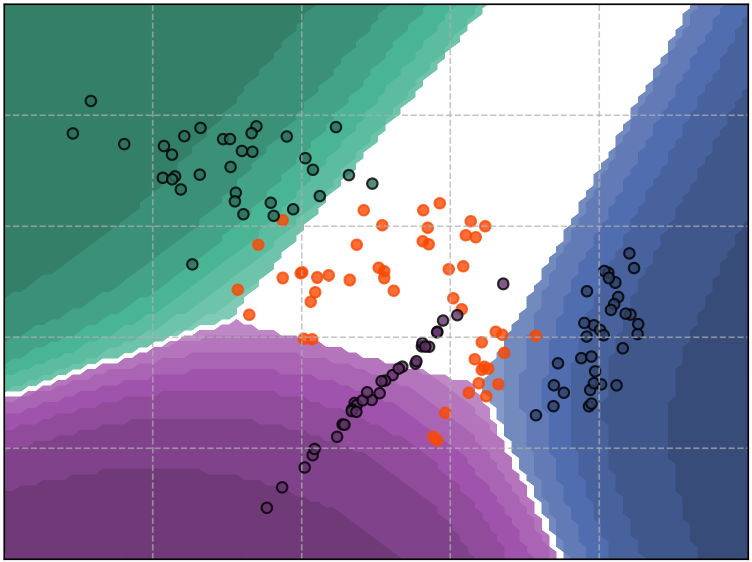} &
		\includegraphics[width=\mytmp\textwidth]{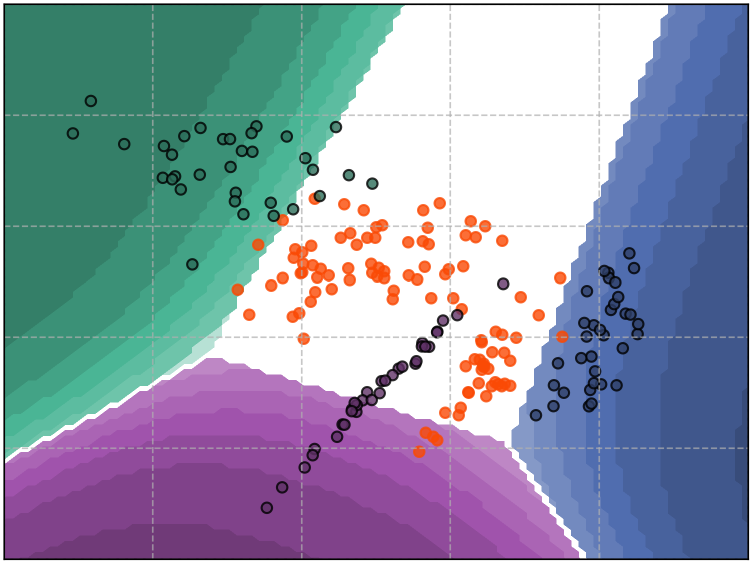} &
		\includegraphics[width=\mytmp\textwidth]{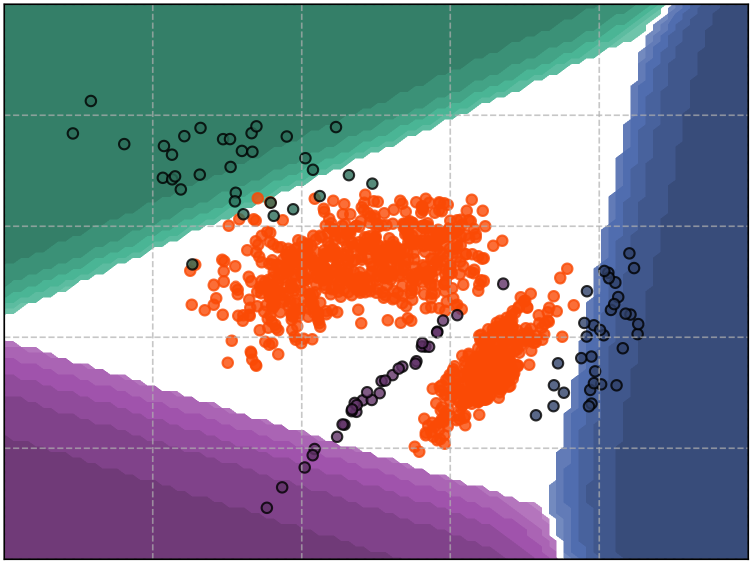} \\
		\textbf{DNN (KvR)} &
		\includegraphics[width=\mytmp\textwidth]{figures/toy-example/ratio/dnn/DNN-ignore-0} &
		\includegraphics[width=\mytmp\textwidth]{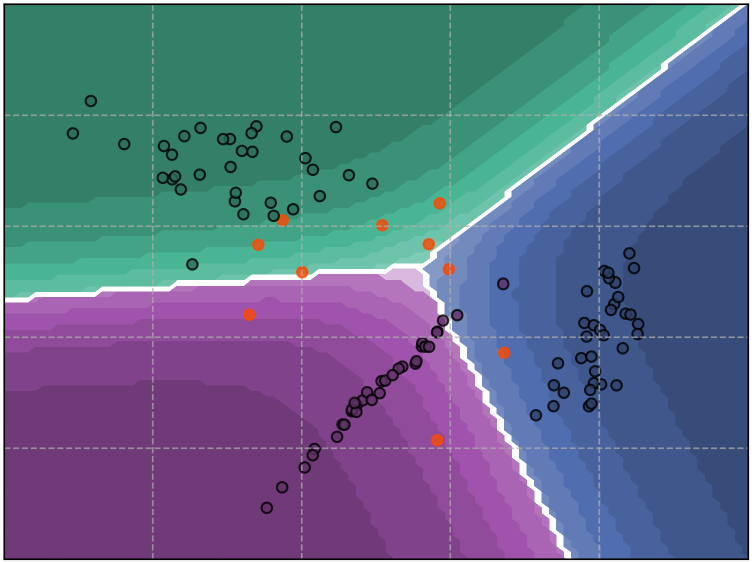} &
		\includegraphics[width=\mytmp\textwidth]{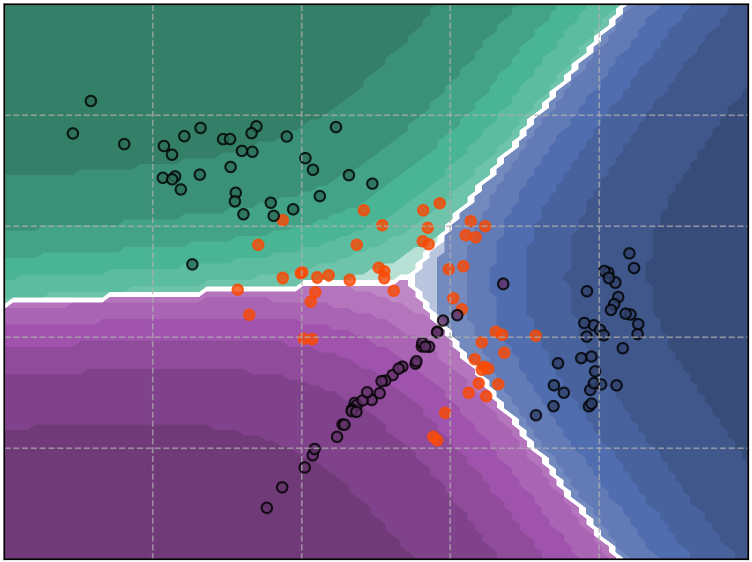} &
		\includegraphics[width=\mytmp\textwidth]{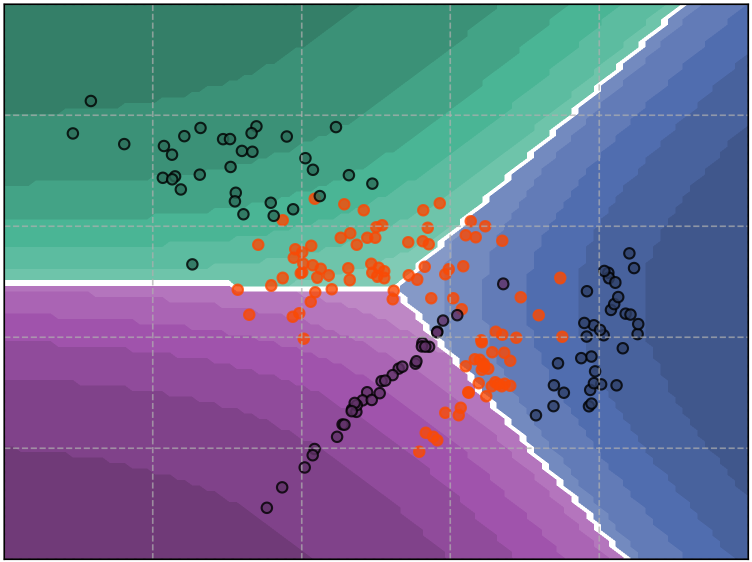} &
		\includegraphics[width=\mytmp\textwidth]{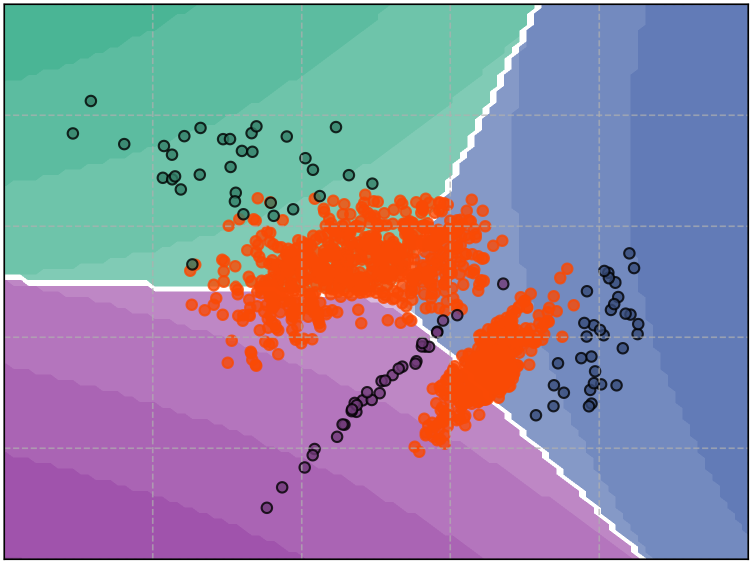} \\
		\bottomrule
	\end{tabular}
\end{table*}%

%% file: tables/evm-cevm-toy-3class.tex
\begin{table*}[tb]
	\centering
	\setlength{\tabcolsep}{1.15pt}
	\renewcommand{\arraystretch}{0.5}
	\renewcommand{\mytmp}{0.173}%
	\caption{EVM (top) and C-EVM (bottom) class boundaries with applied strategies and a column-wise increase in mixup samples.
	This toy dataset contains dark-edged dots from $3$ \acrshortpl{kc} and orange dots as mixups.
	Note that for the C-EVM the visible training and mixup dots are the remaining samples \emph{after} the cluster-based reduction.
	Only in KvR are the mixup samples not reduced.
	Colored areas display class assignment, with opacity indicating confidence, where white is zero confidence or, conversely, high confidence for open space.}\label{tab:evm-cevm-toy-3class}%
	\begin{tabular}{l >{\centering\arraybackslash} m{\mytmp\textwidth} >{\centering\arraybackslash} m{\mytmp\textwidth} >{\centering\arraybackslash} m{\mytmp\textwidth} >{\centering\arraybackslash} m{\mytmp\textwidth} >{\centering\arraybackslash} m{\mytmp\textwidth}}
		\toprule
		\textbf{Strategy} & \multicolumn{5}{c}{\textbf{$\bm{\#}$ \MixKnownRatio}} \\
		\cmidrule{2-6}
		& \num{0} & \num{0.1} & \num{0.5} & \num{1} & \num{10} \\
		\midrule
		\textbf{EVM (SPL)} &
		\includegraphics[width=\mytmp\textwidth]{figures/toy-example/ratio/evm/EVM-ignore-0} &
		\includegraphics[width=\mytmp\textwidth]{figures/toy-example/ratio/evm/EVM-single-10} &
		\includegraphics[width=\mytmp\textwidth]{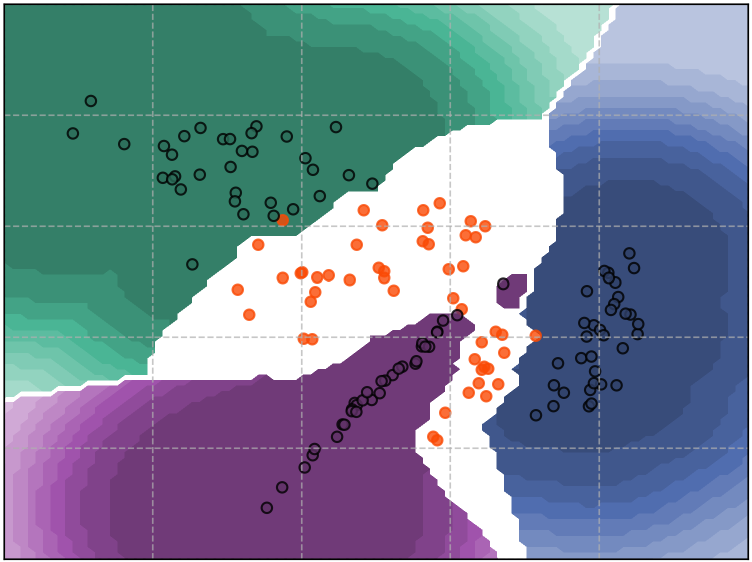} &
		\includegraphics[width=\mytmp\textwidth]{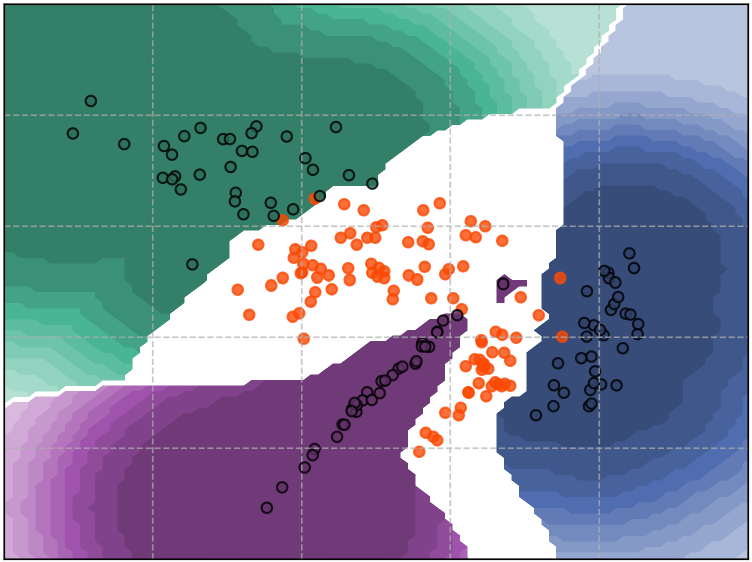} &
		\includegraphics[width=\mytmp\textwidth]{figures/toy-example/ratio/evm/EVM-single-1000} \\
		\textbf{EVM (MPL)} &
		\includegraphics[width=\mytmp\textwidth]{figures/toy-example/ratio/evm/EVM-ignore-0} &
		\includegraphics[width=\mytmp\textwidth]{figures/toy-example/ratio/evm/EVM-multi-10} &
		\includegraphics[width=\mytmp\textwidth]{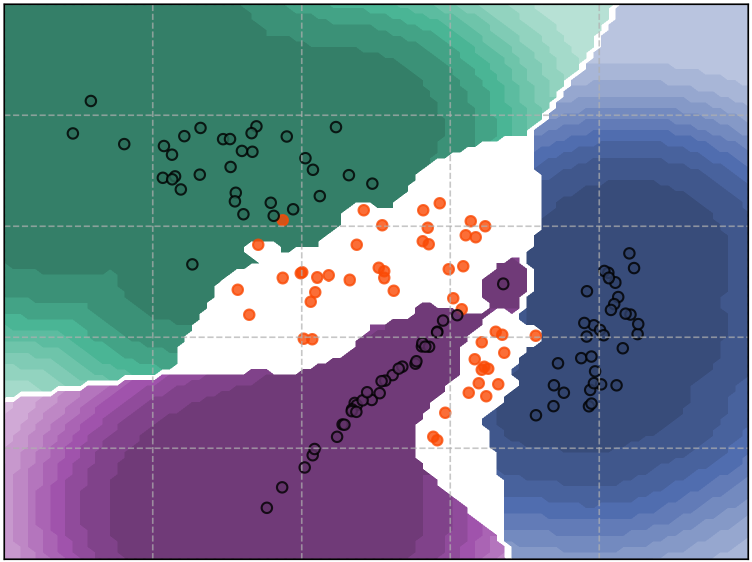} &
		\includegraphics[width=\mytmp\textwidth]{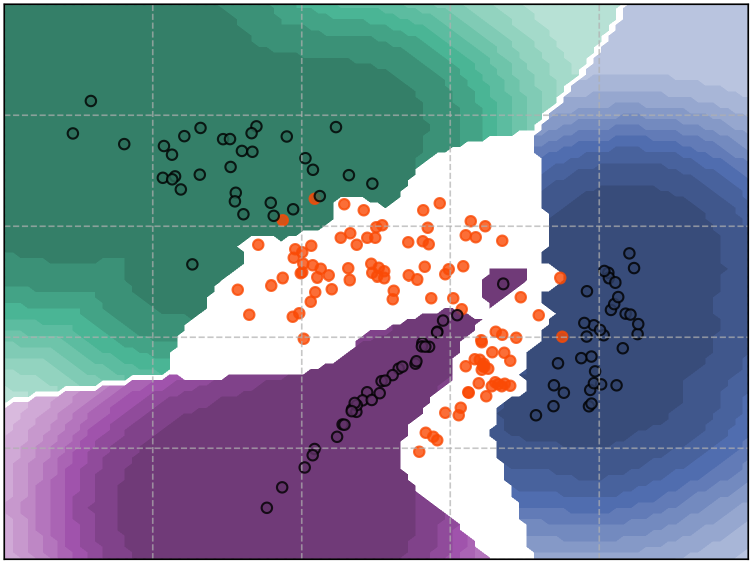} &
		\includegraphics[width=\mytmp\textwidth]{figures/toy-example/ratio/evm/EVM-multi-1000} \\
		\textbf{EVM (KvR)} &
		\includegraphics[width=\mytmp\textwidth]{figures/toy-example/ratio/evm/EVM-ignore-0} &
		\includegraphics[width=\mytmp\textwidth]{figures/toy-example/ratio/evm/EVM-negative-10} &
		\includegraphics[width=\mytmp\textwidth]{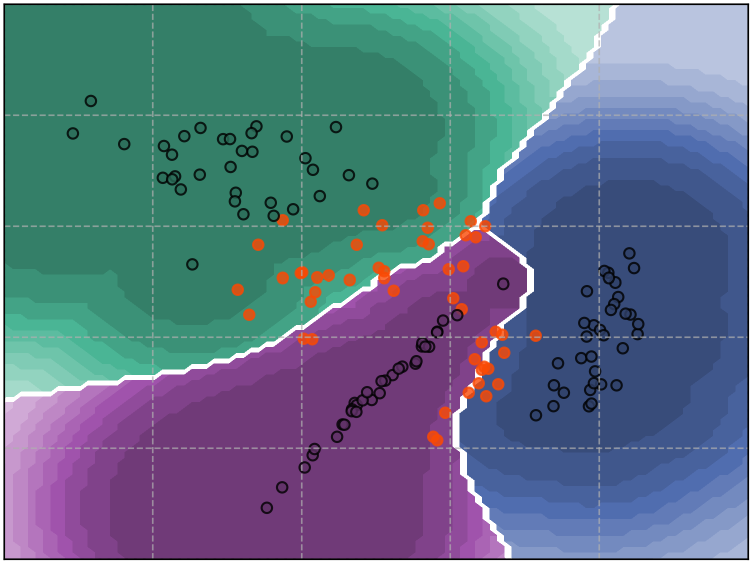} &
		\includegraphics[width=\mytmp\textwidth]{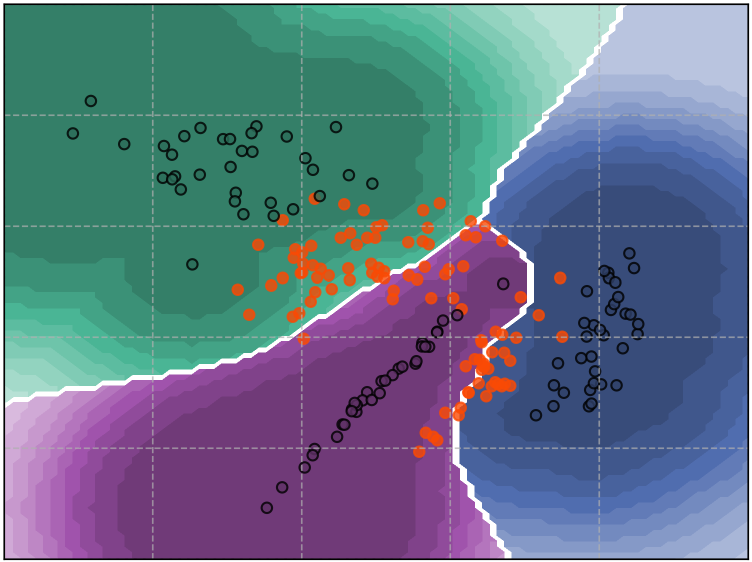} &
		\includegraphics[width=\mytmp\textwidth]{figures/toy-example/ratio/evm/EVM-negative-1000} \\
		\midrule
		\textbf{C-EVM (SPL)} &
		\includegraphics[width=\mytmp\textwidth]{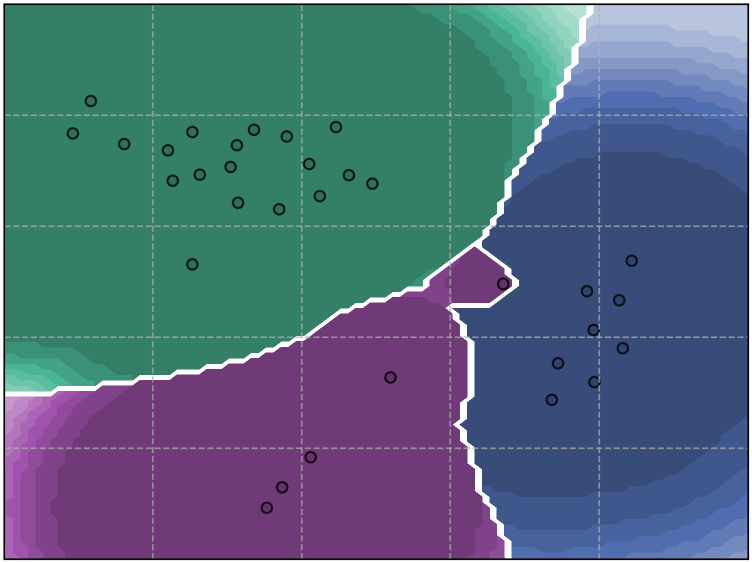} &
		\includegraphics[width=\mytmp\textwidth]{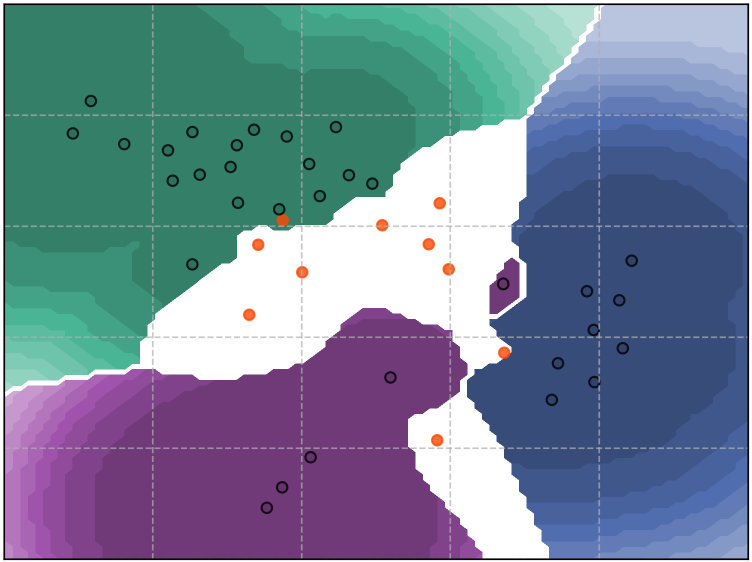} &
		\includegraphics[width=\mytmp\textwidth]{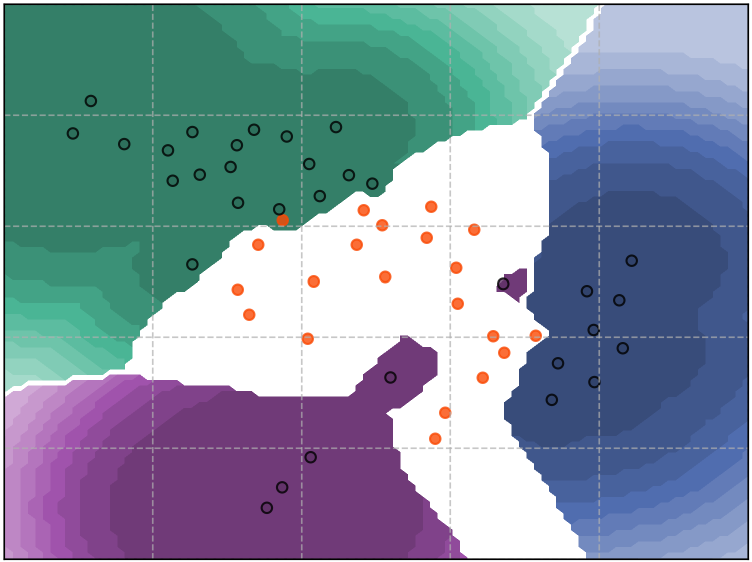} &
		\includegraphics[width=\mytmp\textwidth]{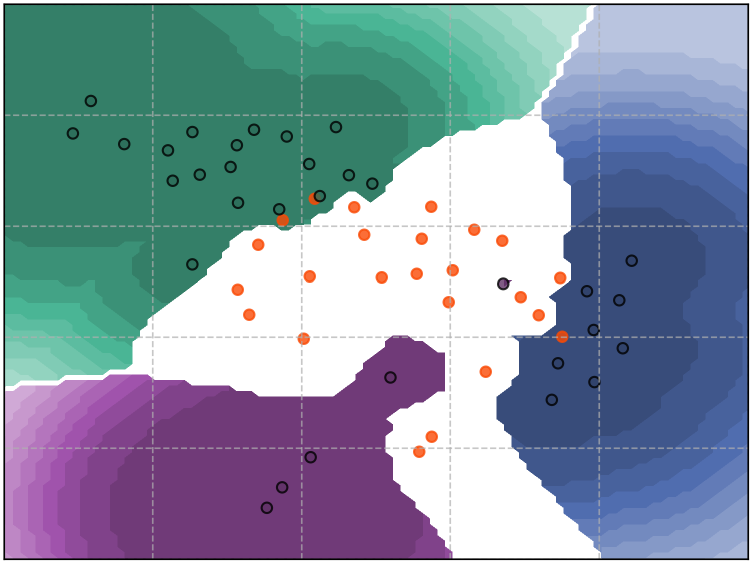} &
		\includegraphics[width=\mytmp\textwidth]{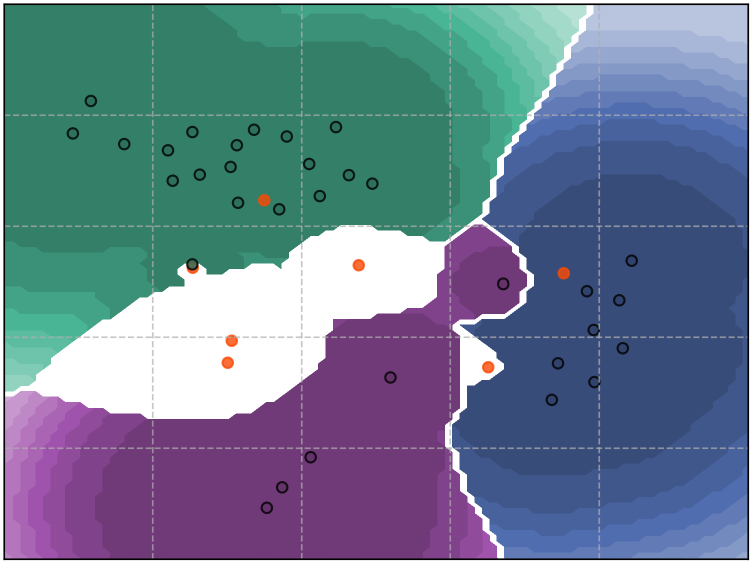} \\
		\textbf{C-EVM (MPL)} &
		\includegraphics[width=\mytmp\textwidth]{figures/toy-example/ratio/cevm/CEVM-ignore-0} &
		\includegraphics[width=\mytmp\textwidth]{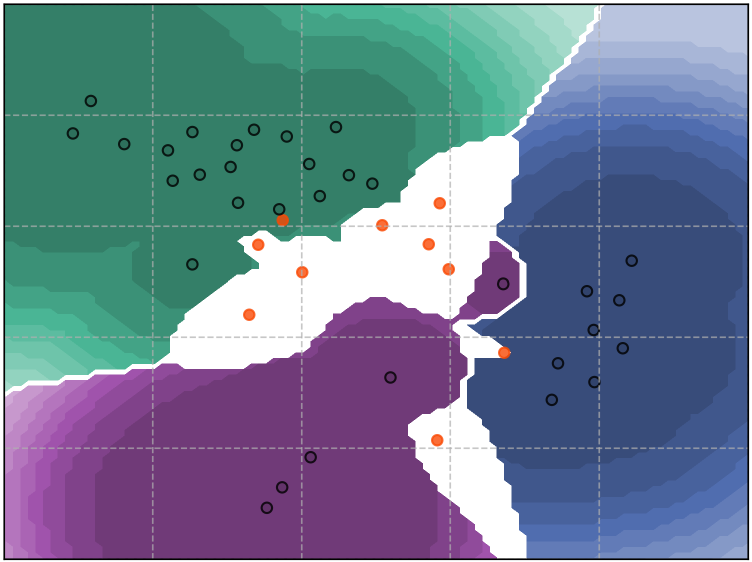} &
		\includegraphics[width=\mytmp\textwidth]{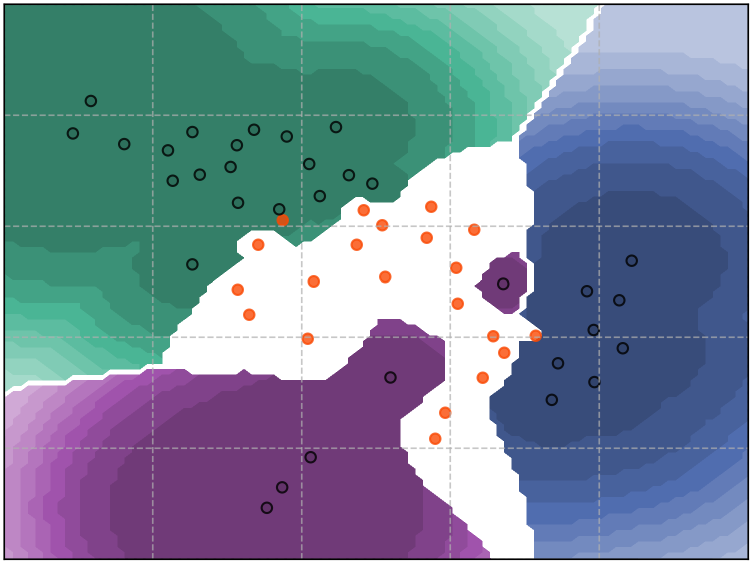} &
		\includegraphics[width=\mytmp\textwidth]{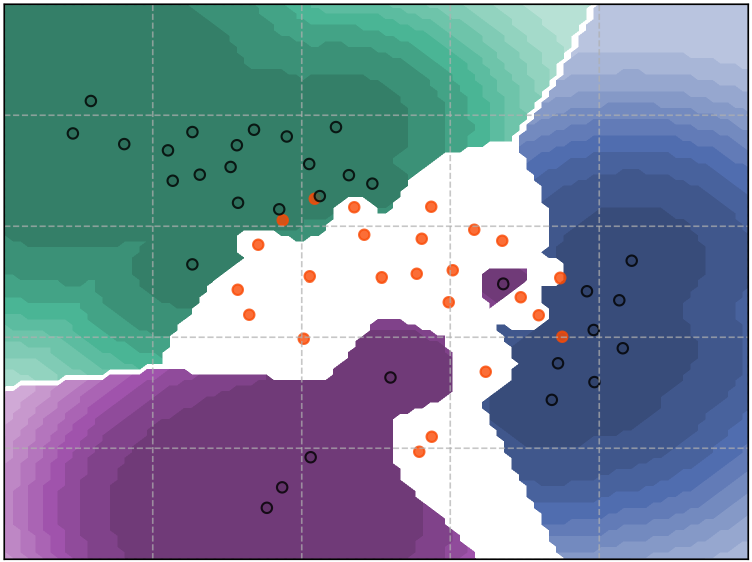} &
		\includegraphics[width=\mytmp\textwidth]{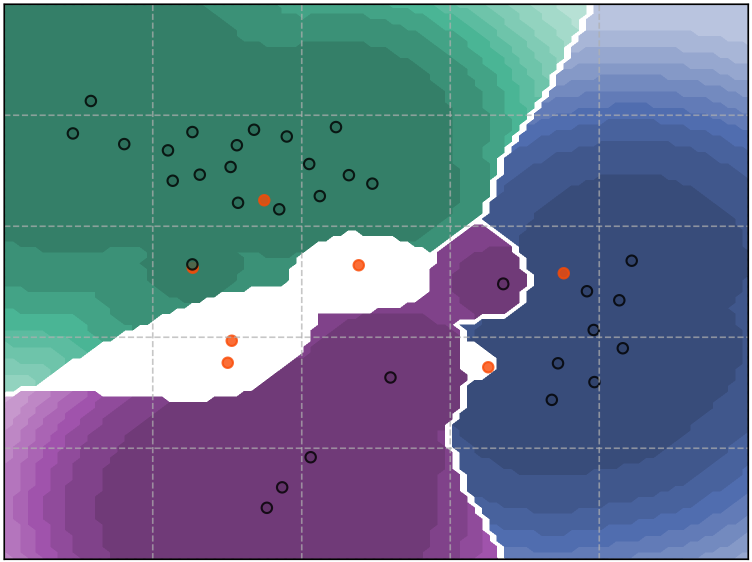} \\
		\textbf{C-EVM (KvR)} &
		\includegraphics[width=\mytmp\textwidth]{figures/toy-example/ratio/cevm/CEVM-ignore-0} &
		\includegraphics[width=\mytmp\textwidth]{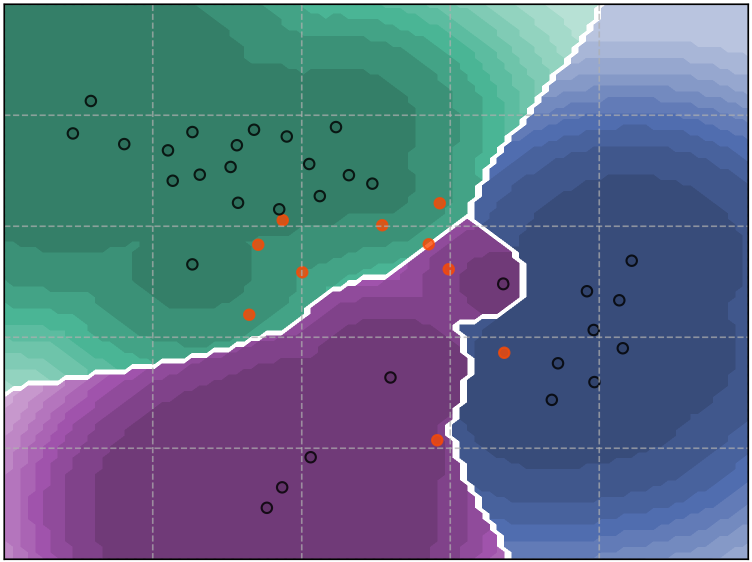} &
		\includegraphics[width=\mytmp\textwidth]{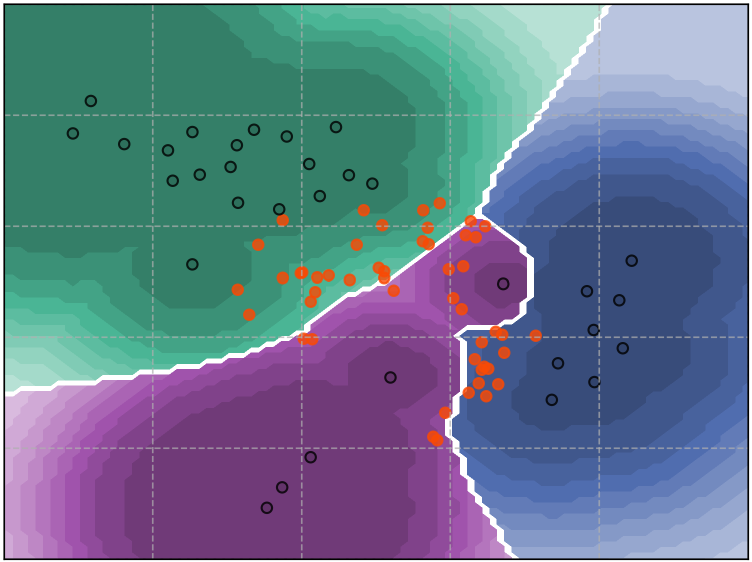} &
		\includegraphics[width=\mytmp\textwidth]{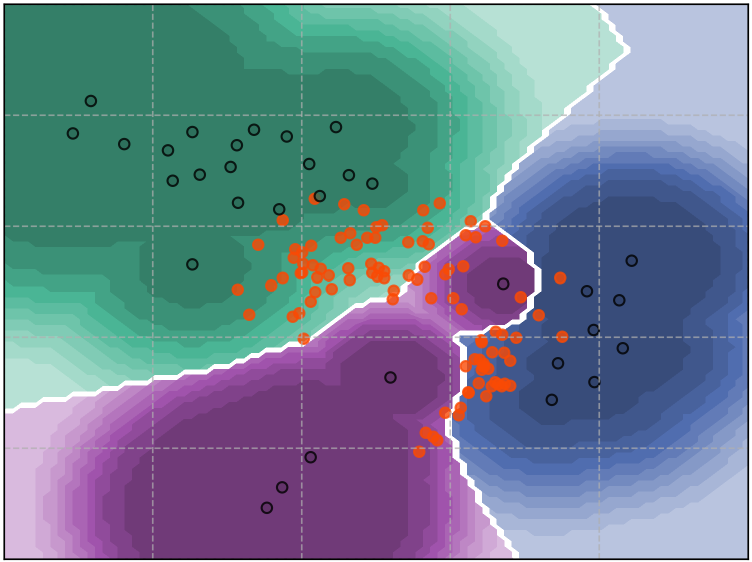} &
		\includegraphics[width=\mytmp\textwidth]{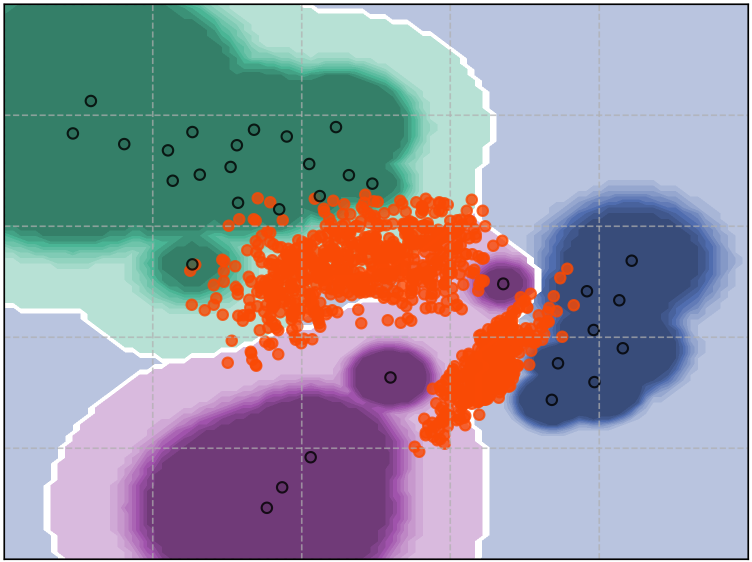} \\
		\bottomrule
	\end{tabular}
\end{table*}%

%% file: tables/svms-toy-3class.tex
\begin{table*}[tb]
	\centering
	\setlength{\tabcolsep}{1pt}
	\renewcommand{\arraystretch}{0.5}
	\renewcommand{\mytmp}{0.17}%
	\caption{W-SVM (top) and \gls{pisvm} (bottom) class boundaries with applied strategies and a column-wise increase in mixup samples.
	This toy dataset contains dark-edged dots from $3$ \acrfullpl{kc} and orange dots as mixups.
	Colored areas display class assignment, with opacity indicating confidence, where white is zero confidence or, conversely, high confidence for open space.}\label{tab:svms-toy-3class}%
	\begin{tabular}{l >{\centering\arraybackslash} m{\mytmp\textwidth} >{\centering\arraybackslash} m{\mytmp\textwidth} >{\centering\arraybackslash} m{\mytmp\textwidth} >{\centering\arraybackslash} m{\mytmp\textwidth} >{\centering\arraybackslash} m{\mytmp\textwidth}}
		\toprule
		\textbf{Strategy} & \multicolumn{5}{c}{\textbf{$\bm{\#}$ \MixKnownRatio}} \\
		\cmidrule{2-6}
		& \num{0} & \num{0.1} & \num{0.5} & \num{1} & \num{10} \\
		\midrule
		\textbf{W-SVM (SPL)} &
		\includegraphics[width=\mytmp\textwidth]{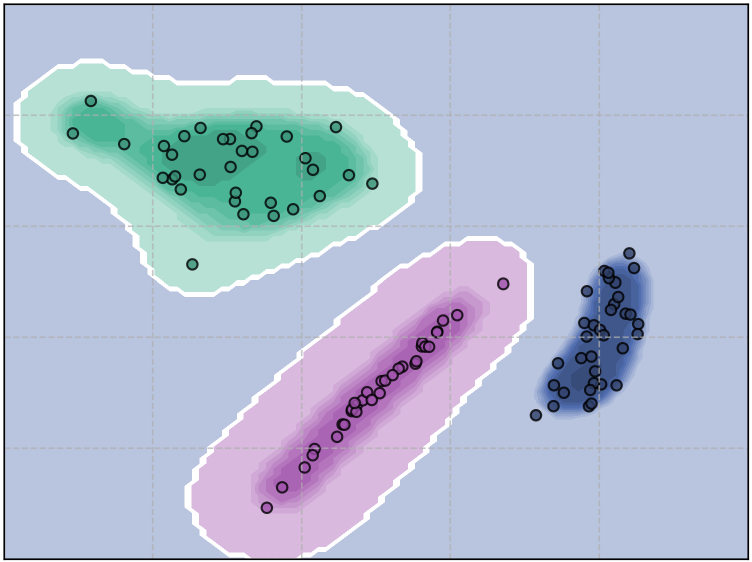} &
		\includegraphics[width=\mytmp\textwidth]{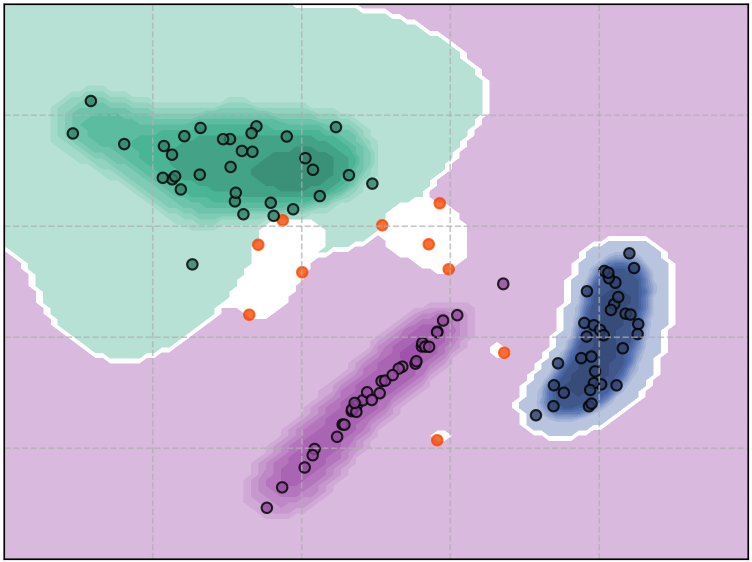} &
		\includegraphics[width=\mytmp\textwidth]{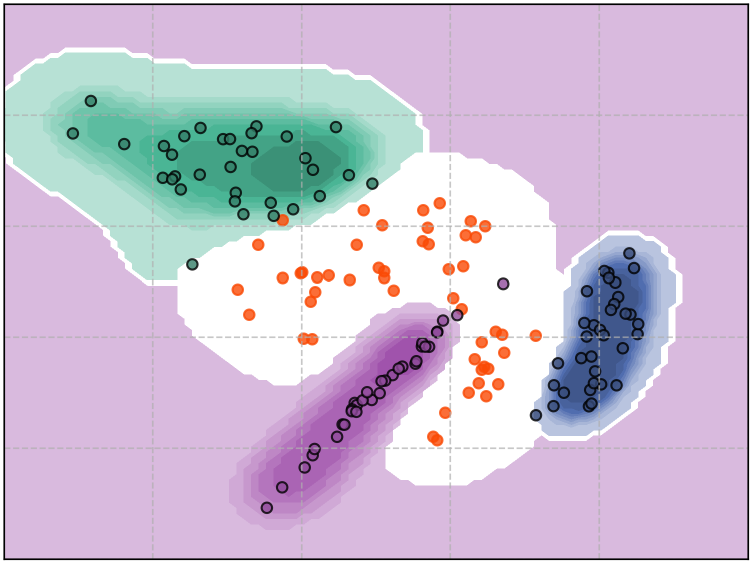} &
		\includegraphics[width=\mytmp\textwidth]{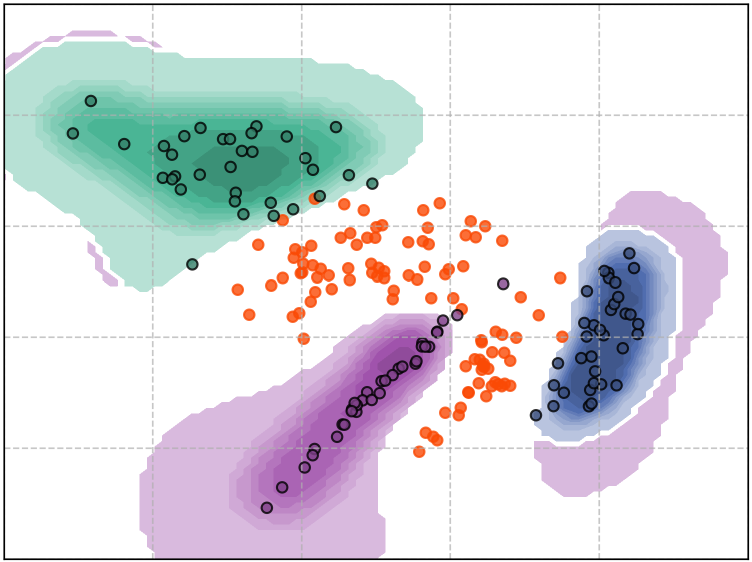} &
		\includegraphics[width=\mytmp\textwidth]{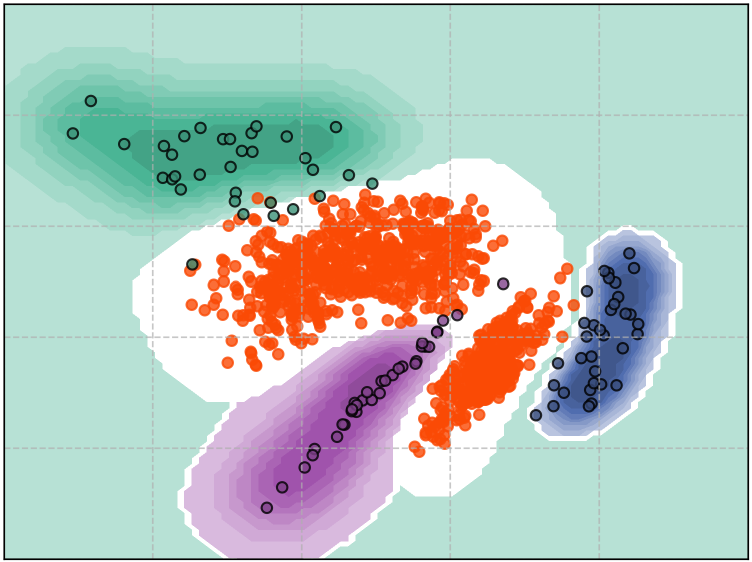} \\
		\textbf{W-SVM (KvR)} &
		\includegraphics[width=\mytmp\textwidth]{figures/toy-example/ratio/wsvm/W-SVM-ignore-0} &
		\includegraphics[width=\mytmp\textwidth]{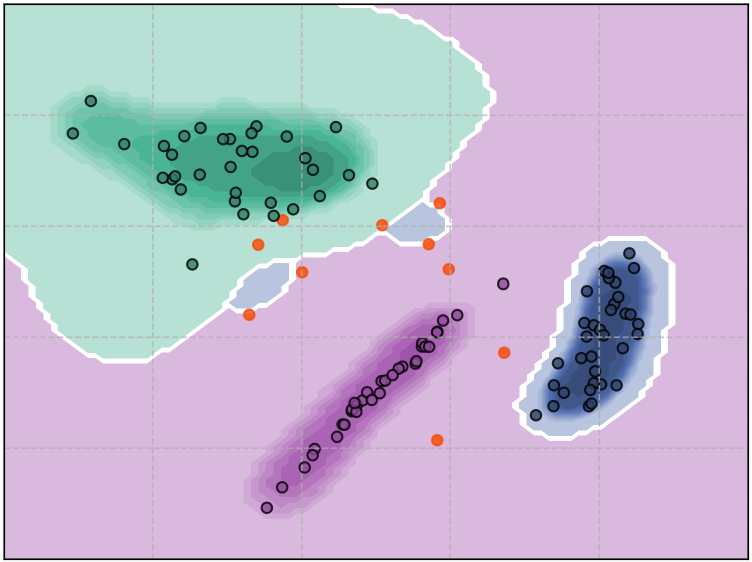} &
		\includegraphics[width=\mytmp\textwidth]{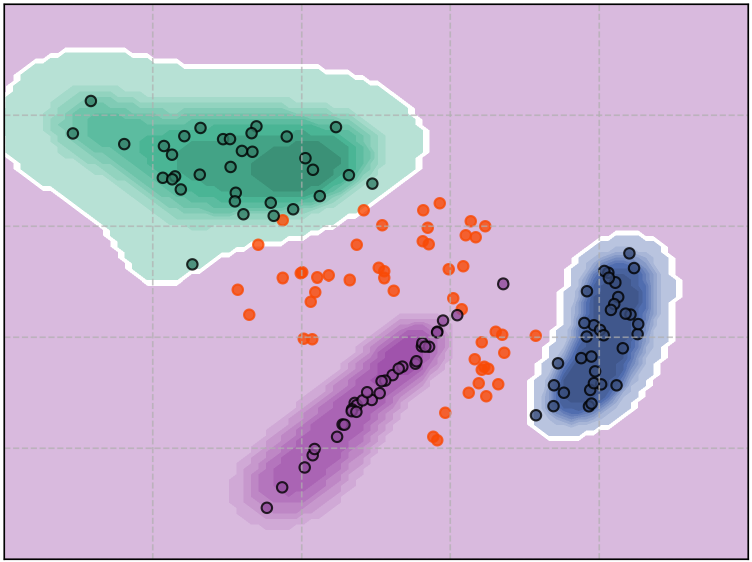} &
		\includegraphics[width=\mytmp\textwidth]{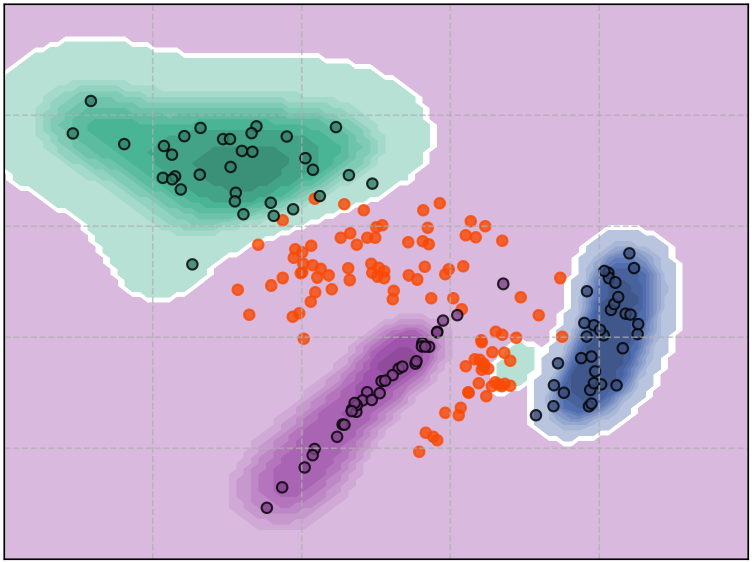} &
		\includegraphics[width=\mytmp\textwidth]{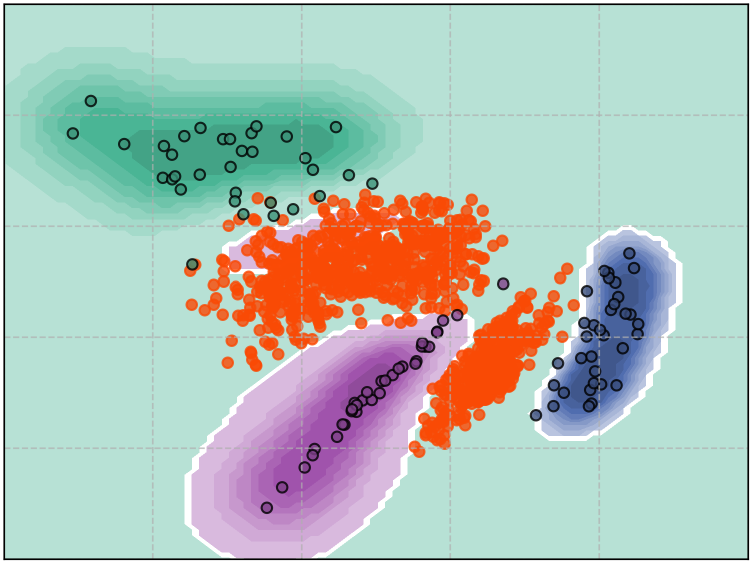} \\
		\midrule
		\textbf{\acrshort{pisvm} (SPL)} &
		\includegraphics[width=\mytmp\textwidth]{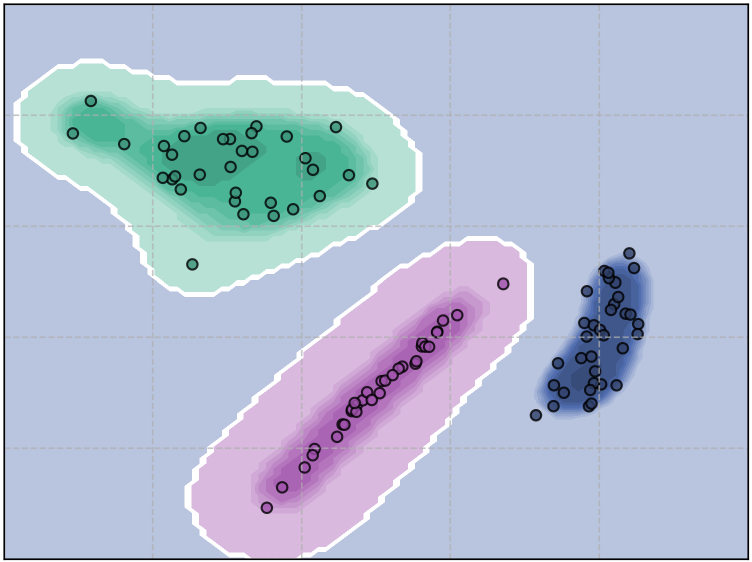} &
		\includegraphics[width=\mytmp\textwidth]{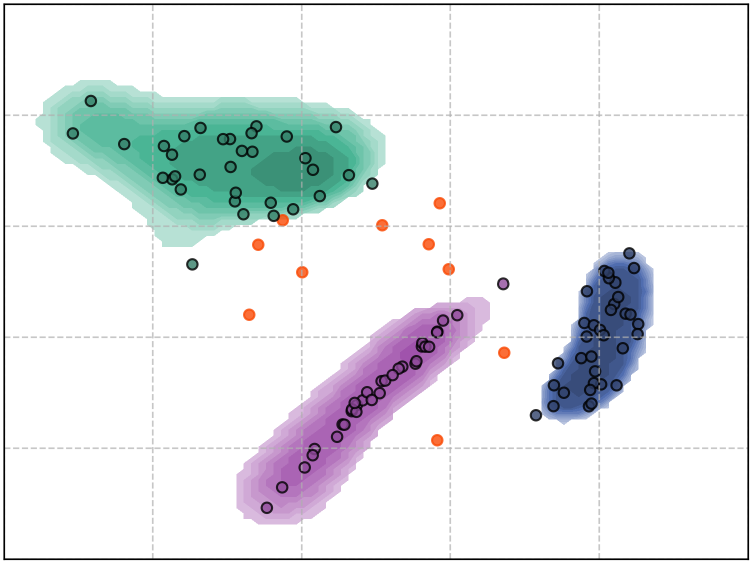} &
		\includegraphics[width=\mytmp\textwidth]{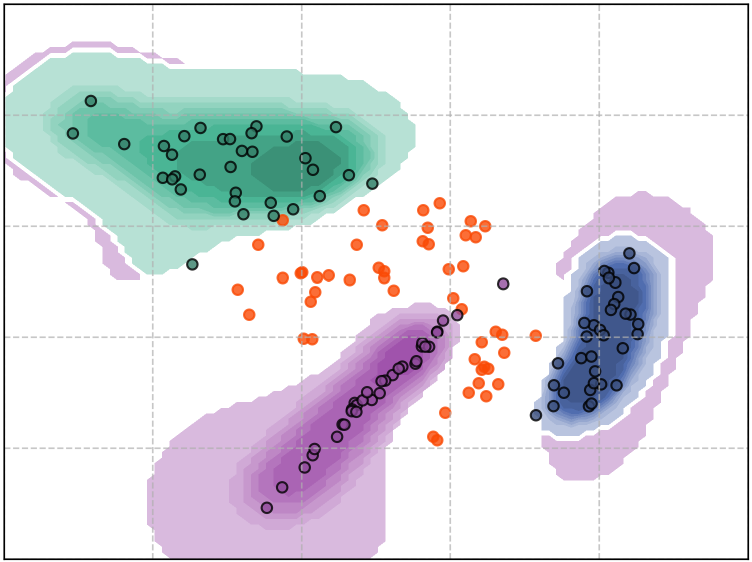} &
		\includegraphics[width=\mytmp\textwidth]{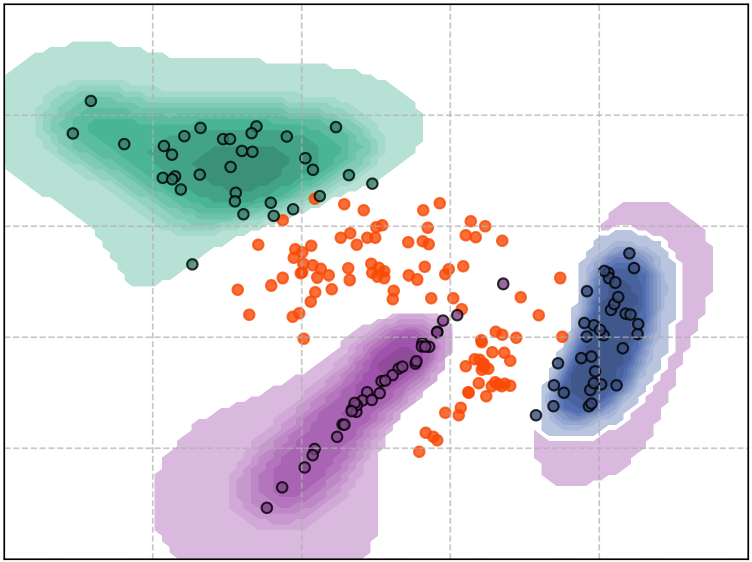} &
		\includegraphics[width=\mytmp\textwidth]{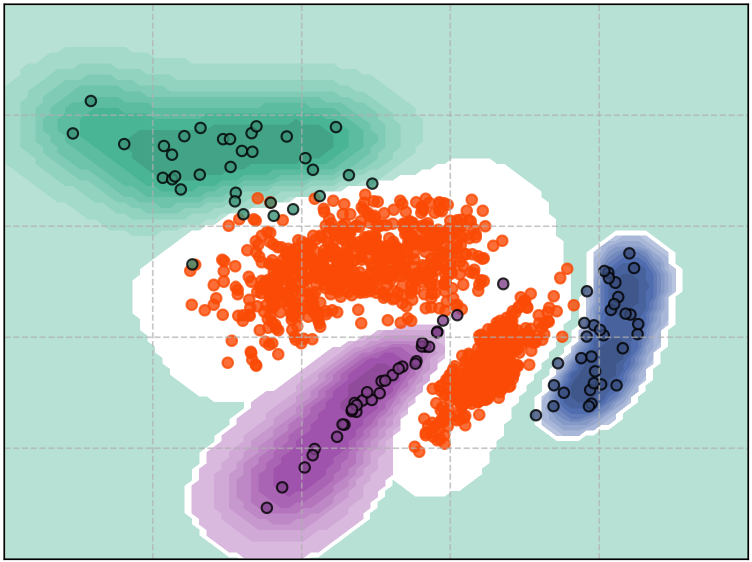} \\
		\textbf{\acrshort{pisvm} (KvR)} &
		\includegraphics[width=\mytmp\textwidth]{figures/toy-example/ratio/pisvm/PI-SVM-ignore-0} &
		\includegraphics[width=\mytmp\textwidth]{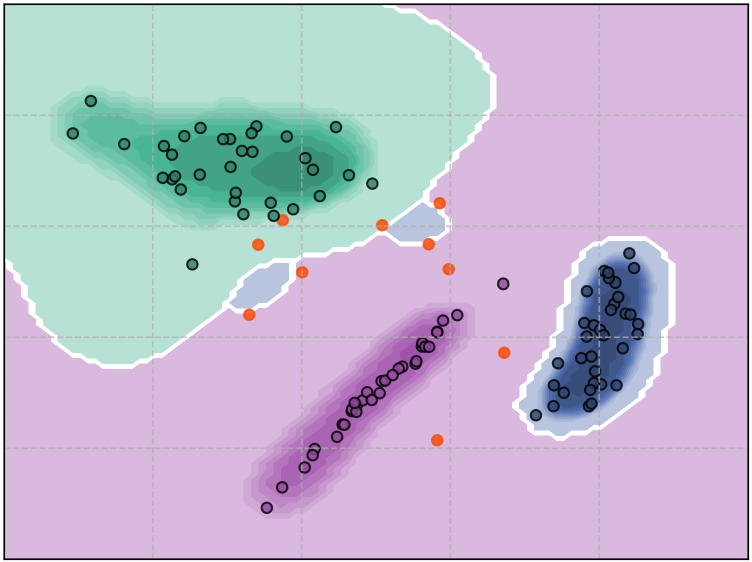} &
		\includegraphics[width=\mytmp\textwidth]{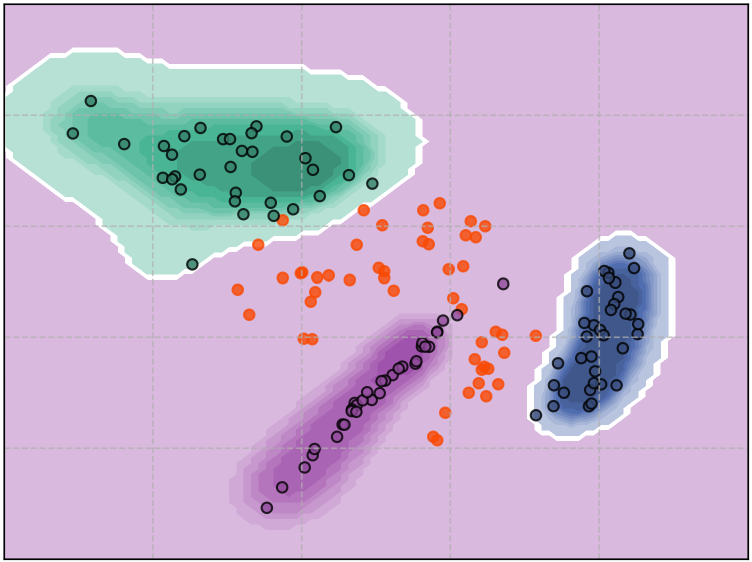} &
		\includegraphics[width=\mytmp\textwidth]{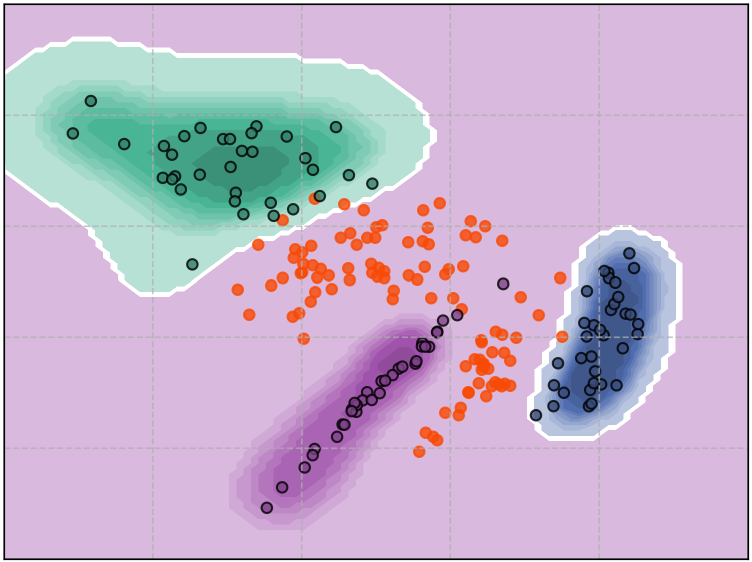} &
		\includegraphics[width=\mytmp\textwidth]{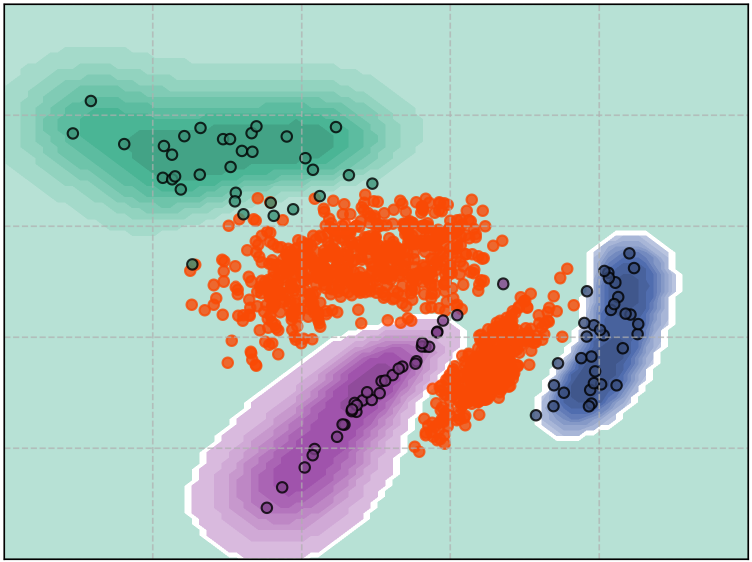} \\
		\bottomrule
	\end{tabular}
\end{table*}%